\documentclass[final,5p,times,twocolumn]{elsarticle}

\setcitestyle{authoryear,round,semicolon}

\def\hideInternalStuff{1} 
\def\includeOldVersions{0} 

\PassOptionsToPackage{hyphens}{url}
\usepackage{hyperref}

\usepackage{bm}
\usepackage{amsmath,amssymb,amsfonts}
\usepackage{algorithmic}
\usepackage{graphicx}
\graphicspath{{./figures/}
}
\usepackage{textcomp}
\usepackage{}
\usepackage{url}
\usepackage{graphbox}
\usepackage{xcolor}
\usepackage{subcaption}

\newcommand{\todo}[1]{\color{blue}[TODO: {#1}]\normalcolor}
\definecolor{noteToSelfColor}{rgb}{0.6,0.6,0.6}
\newcommand{\noteToSelf}[1]{\color{noteToSelfColor}[{#1}]\normalcolor}
\definecolor{todoDoneColor}{rgb}{0.0,0.6,0.0}
\newcommand{\todoDone}[1]{\color{todoDoneColor}[Done: {#1}]\normalcolor}
\definecolor{unresolvedColor}{rgb}{0.6,0.0,0.0}

\definecolor{discussedWithChiaraColor}{rgb}{1.0,0.65,0.0}
\newcommand{\discussedWithChiara}[1]{\color{discussedWithChiaraColor}[{#1}]\normalcolor}

\definecolor{questionColor}{rgb}{1.0,1.0,0.0}

\definecolor{answerColor}{rgb}{0.0,1.0,0.0}
\newcommand{\answer}[1]{%
  \colorbox{answerColor}{%
    \begin{minipage}[t]{\linewidth}%
        {ANSWER: #1}%
     \end{minipage}%
  }%
}

\definecolor{textSnippetColor}{rgb}{0.6,0.6,0.6}
\newcommand{\textSnippet}[1]{\color{textSnippetColor}[\textit{#1}]\normalcolor}
\definecolor{revisionColor}{rgb}{0.0, 0.0, 0.0}
\newcommand{\revision}[1]{\color{revisionColor}{#1}\normalcolor}
\definecolor{revision2Color}{rgb}{0.0, 0.0, 0.0}
\newcommand{\revisiontwo}[1]{\color{revision2Color}{#1}\normalcolor}

\newcommand{\fat}[1]{\mathbf{#1}} 
\newcommand{\bldgr}[1]{\boldsymbol{#1}} 

\newlength{\myWidth}

\newcommand{\myclearpage}{\clearpage}

\usepackage{natbib}

\if\hideInternalStuff1
  \renewcommand{\todoDone}[1]{}
  \renewcommand{\noteToSelf}[1]{}
  \renewcommand{\answer}[1]{}
  \renewcommand{\textSnippet}[1]{}
  \renewcommand{\myclearpage}{}
  \renewcommand{\discussedWithChiara}[1]{}
\fi

\journal{Medical Image Analysis}

\begin{document}

\begin{frontmatter}



\title{A Lightweight 
\revision{Generative} 
Model for Interpretable 
Subject-level
Prediction}



\author[1,2]{Chiara Mauri\corref{cor1}}
  \ead{cmauri@mgh.harvard.edu}
\author[2]{Stefano Cerri}
\author[3]{Oula Puonti} 
\author[4]{Mark M\"{u}hlau} 
\author[2,5]{Koen Van Leemput}

\cortext[cor1]{Corresponding author}
%
\address[1]{Department of Health Technology, Technical University of Denmark,
Denmark}

\address[2]{Athinoula A. Martinos Center for Biomedical Imaging, Massachusetts General Hospital, Harvard Medical School, 
USA}           

\address[3]{Danish Research Centre for Magnetic Resonance, Center for Functional and Diagnostic Imaging and Research, Copenhagen University Hospital Hvidovre, 
Denmark}

\address[4]{Department of Neurology and TUM-Neuroimaging Center, School of Medicine, Technical University of Munich, 
Germany}

\address[5]{Department of Neuroscience and Biomedical Engineering, Aalto University,
Finland}



\begin{abstract}
Recent years have seen a growing interest in methods for predicting 
\revision{an unknown}
variable of interest, such as a subject's diagnosis, from 
medical images
\revision{depicting its anatomical-functional effects}%
.
Methods based on discriminative modeling excel at making accurate predictions, but are challenged in their ability to explain their decisions in 
anatomically
meaningful terms.
In this paper, we propose a simple technique for single-subject prediction that is inherently interpretable.
It augments the generative models used in classical human brain mapping techniques, in which 
\revision{the underlying}
cause-effect relations can be encoded, with a multivariate noise model that captures dominant spatial correlations.
Experiments demonstrate that the resulting 
model can be efficiently inverted to make accurate subject-level predictions, while at the same time offering intuitive 
\revision{visual}
explanations of its inner workings. 
The method is easy to use: training is fast for typical training set sizes,
and 
only a single hyperparameter needs to be set by the user.
%
Our code
is available at \url{https://github.com/chiara-mauri/Interpretable-subject-level-prediction}. 
\end{abstract}



\begin{keyword}
Image-based prediction \sep 
Brain age \sep
Explainable AI \sep
Generative models


\end{keyword}

\end{frontmatter}



\section{Introduction}
\label{sec:introduction}

\noteToSelf{
Intro about prediction. Ideally not only predict well, but also provide insight into what is driving the predictions. Current methods can't do both.
}

\noindent
\revision{%
Single-subject prediction  
methods aim to infer 
a subject's
underlying clinical condition
-- such as their disease status --
from 
its observed effect
on the subject's anatomy or function
as 
measured
by medical imaging. 
%
%
%
%
The ability to perform this task accurately
would have%
}
numerous potential applications in diagnosing disease, tracking progression, and evaluating treatment.
It could
also
help clinicians to prospectively identify which patients are at highest risk of future disability accrual,
leading to better counseling of patients and better overall clinical outcomes.

%
Many methods for automatic single-subject prediction have  
been proposed in the literature, 
using multivariate techniques that 
combine
the weakly predictive power of many voxels simultaneously to obtain accurate predictions at the subject level%
~\citep{arbabshirani2017single,cole2019quantification}. Recent years have seen 
a 
rapid 
growth in
methods for predicting a subject's \emph{age} 
from their
brain scan, 
in particular, 
with the gap between 
the estimated and the real 
age 
being suggested as a potential biomarker of 
neurological disease~\citep{cole2019quantification,kaufmann2019common}.
Although 
high prediction accuracies can now be achieved,
especially when 
methods are trained on the very large datasets that have recently become available~\citep{german2014german,breteler2014ic,schram2014maastricht,miller2016multimodal,alfaro2018image},
comparatively little attention has been paid to 
\emph{interpretability},
i.e., to the ability to explain the predictions 
to clinicians in terms 
that are biologically 
meaningful.
%
Nevertheless, such interpretability is likely required before automated prediction methods can safely be adopted for widespread clinical use~\citep{rudin2019stop}.

\noteToSelf{
Discriminative methods. List a few classical techniques (all linear?), then DL. [ Here or intro paragraph above: perhaps mention age prediction specifically ] However, struggle to provide insight. Post-hoc explanation methods -- such as saliency maps plus their definition as measuring gradient of decision surface [ or maybe better how prediction changes as input is varied ] -- have been criticized. Even linear -- simple saliency maps -- have been criticized as they show \emph{what} but not \emph{why}. [ cite Haufe: zeroes becomes nonzero, high weights don't mean anything
}
\noindent

A key difficulty in obtaining interpretability is that almost all
\revision{subject-level}
prediction methods 
are currently based on \emph{discriminative} 
learning, in which 
a direct mapping 
from an input image 
to a variable of interest 
is 
estimated
from 
training 
examples. 
%
%
%
Especially with the deep neural networks that have become prominent in recent years,
this results in ``black box'' models 
whose internal workings 
are hard to 
explain to
humans.
Although many of the post hoc explanation methods
\citep{Ras2022JAIR,arrieta2020explainable,Baehrens2010JMLR,Sundararajan2017ICML,
Springenberg2014arxiv,Selvaraju2017ICCV,smilkov2017smoothgrad,zeiler2014visualizing,bach2015pixel}
%
that have been developed for 
such 
complex
models
have 
\todoDone{Mark says: what do you mean, ``specific issues of their own''?}
raised specific criticism%
~\citep{Arun2021RAI,Ghassemi2021Lancet,adebayo2018sanity,rudin2019stop, wilming2022scrutinizing,sixt2020explanations,gu2019saliency},
a more fundamental challenge is that
interpretability 
is 
\emph{intrinsically}
hard 
for discriminative 
\revision{subject-level prediction}
methods 
--
even when
very simple
(e.g., linear)
models are used.
This is because 
discriminative  
methods optimize their prediction performance not only by amplifying the signal of interest in the data, but also 
by suppressing unrelated 
``distractor''
patterns,
so that their reason for looking at specific voxels cannot easily be 
deduced \revision{\citep{haufe2014interpretation,wilming2022scrutinizing,weichwald2015causal}}.
%
%
Therefore, 
while e.g., linear discriminative methods 
are
trivially 
transparent 
about
\emph{how}
they compute their
results 
(the weight they give to each image area can readily be inspected),
they offer no explanation of \emph{why} 
they are using specific image areas more than others \citep{Ghassemi2021Lancet,rudin2019stop,haufe2014interpretation,wilming2022scrutinizing,lipton2018mythos}. 
Stated differently,
their explanation refers to an understanding of how the model works, as opposed to an explanation of how the world works~\citep{rudin2019stop}.

\noteToSelf{
Generative methods. Brain mapping, group level. Show intuitive explanations, even causal effects in some cases (e.g., age and sex). However, unlike discriminative methods (which as multivariate combine power of many/all voxels simultaneously) mass-univariate, so consider each voxel independently, each too weak to predict on the subject-level.
}
\noindent
  
%
Classical human brain mapping techniques, 
originally developed for analyzing functional images~\citep{friston1991comparing,worsley1992three,friston1994statistical,worsley1995analysis} but later adapted for structural imaging~\citep{chung2001unified,davatzikos2001voxel,wright1995voxel,ashburner2000voxel,fischl2000measuring,snook2007voxel},
are based on \emph{generative}
rather than on \emph{discriminative} models:
They encode a mapping from a variable of interest to the image domain, rather than vice versa.
Their aim is to identify, on a population level, brain regions that are significantly
correlated with specific variables of interest
(such as disease status or an experimental condition)
 or their interactions.
Especially 
when the variables of interest have a \emph{causal} effect on brain anatomy
-- 
as is the case for e.g., age, gender or a particular brain disease
--
these methods are inherently interpretable%
:
The spatial maps they provide indicate how 
each brain location 
would change, on average, if we had the ability to control the variable of interest at will%
\footnote{Assuming the absence of uncontrolled confounding variables, 
\revision{see Fig.~\ref{fig:graphs}}.
}%
.
%
However,
because
classical
brain mapping techniques only consider each voxel-level measurement \emph{independently} (so-called mass-univariate modeling), they are unsuitable for subject-level prediction,
as each 
individual
voxel 
by itself
is 
typically
only weakly predictive of the variable of interest.

\noteToSelf{
Here contribution: augment/endow brain mapping technique -- modeling causal effect on a group-level -- with [ low dimensional latent ] noise model that captures the domain correlations in the data. This can be efficiently [ minutes CPU time for a few hundred subjects, as in many practical scenarios ] inverted to obtain accurate subject-level predictions. Easy to use -- single hyperparameter -- and fast [ if not said before, here minutes CPU time for a few hundred subjects, as in many practical scenarios ]. 
[ Maybe here inherently interpretable because causal: group-level and subject-level (counterfactuals) ] 
We show experiments demonstrating competitive prediction performance.
}
\noindent

In this paper, 
we propose 
a new 
method 
that 
aims to
combine 
the 
predictive 
power
of 
state-of-the-art
multivariate models
on the one hand,
with 
the 
superior
interpretability properties 
of 
classical 
brain mapping techniques
on the other.
%
%
This is accomplished by 
generalizing
the 
independent,
voxel-wise
noise model
in 
those
classical 
techniques
with 
one
that 
also 
takes into account
correlations 
\emph{between} voxels.
%
%
In particular,
we use a 
linear-Gaussian latent variable 
model
that allows us
to
simultaneously
capture the dominant spatial correlations in the noise,
control the number of free parameters that need to be learned from training data,
and 
efficiently ``invert'' the model to make accurate subject-level predictions.
Because it inherits the cause-effect relation modeling from classical brain mapping techniques,
the approach is inherently interpretable:
It provides 
both 
population-level spatial maps 
of
the average effect of variables of interest on brain shape,
as well as
the ability to 
apply
these
effects at the level of 
individual subjects (so-called counterfactuals).
%
The method is 
flexible and easy to use:
It has only a single hyperpameter that needs to be tuned by the user,
training is typically fast even 
without hardware acceleration,
and incorporating additional covariates 
(and their possible interactions with the variable of interest)
is straightforward.

\noteToSelf{
[ Structure of the paper is as follows. ]
Earlier version appeared in~\citep{mauri2022accurate},
}
\noindent


An early version of this work appeared in~\citep{mauri2022accurate}.
Here we significantly expand on the basic algorithm described in that paper,
conducting an in-depth analysis of both the interpretability and 
the prediction performance of the method,
detailing a fast practical implementation,
and 
introducing extensions where dependencies on the variables of interest are nonlinear.

%

\noteToSelf{
Things that are different from the MICCAI paper:
\begin{list}{-}{}

\item Practical implementation

\item Bias-variance trade-off

\item Interpretability analysis [ incl. age-specific templates and counterfactuals, dependency of $\fat{w}_D$ on $N$ ]

\item Extensions: nonlinear dependencies on the variable of interest; additional subject-specific covariates

\end{list}
}

\revision{
\section{Context and related work}

\subsection{Problem statement}

%
%

\noindent
Fig.~\ref{fig:graphs}(a) illustrates the problem we 
address in this paper in its most basic form.
An unknown quantity $x$ in a 
patient 
is causing anatomical changes in their brain scan $\fat{t}$.
By modeling the causal relationship between $x$ and $\fat{t}$, we aim both to estimate the unknown value of $x$ from $\fat{t}$ and to provide intuitive explanations of the estimation procedure -- for instance by synthesizing images at different values of $x$ and letting the user visually compare those with $\fat{t}$
(see Fig.~\ref{fig:counterfactuals} for examples).
%
Models of the type shown in Fig.~\ref{fig:graphs}(a) are known as \emph{generative} models in the machine learning community, or as \emph{encoding} models in the neuroimaging literature~\citep{friston2008bayesian,weichwald2015causal}.

For the purpose of estimating the parameters of the model, we need access to
a set of training images for which the underlying value of $x$ is known.
It is worth mentioning that,
unlike in age-prediction experiments 
where there is 
an abundance of available training data,
many 
scenarios encountered in practice
will provide only a few thousand training subjects at best:}
%
%
Imaging datasets collected for studying specific diseases 
typically have  at most 1,000-3,000 subjects%
~\citep{jack2008alzheimer,di2014autism,ellis2009australian,satterthwaite2014neuroimaging},
whereas even the largest prospective cohort imaging studies~\citep{german2014german,breteler2014ic,schram2014maastricht,miller2016multimodal,alfaro2018image}
contain only a modest number of subjects with specific diseases
(e.g., 
of the 
100,000 participants projected to be scanned in the UK Biobank~\citep{alfaro2018image},
only 
around 200 and 1,000 can be expected to be multiple sclerosis and epilepsy patients,
respectively~\citep{mackenzie2014incidence,epilepsycouncil2011}).
%

\revision{
As we shall see, it is straightforward to generalize the model of Fig.~\ref{fig:graphs}(a) to additionally include known subject-level covariates $\fat{y}$   
that 
cause their own anatomical changes in $\fat{t}$
(independent of $x$).
The resulting model,
which is illustrated in Fig.~\ref{fig:graphs}(b),
can be used to try and help improve prediction accuracy.
%
When 
covariates are 
(partial)
causes of $x$ itself,
as illustrated in Fig.~\ref{fig:graphs}(c),
they are known as \emph{confounders} --
common underlying causes of both $x$ and $\fat{t}$.
It is well-known~\citep{pearl2018book} that 
confounders should 
be included in the model to 
preserve 
its
ability to capture the causal effect of $x$ on $\fat{t}$,
as will be illustrated in Sec.~\ref{sec:covariates}.

%
%
%
%
%
%

\begin{figure}
  \setlength{\myWidth}{\linewidth}
  \setlength{\fboxrule}{2pt}
  \fcolorbox{white}{white}{%
  \includegraphics[width=\myWidth]{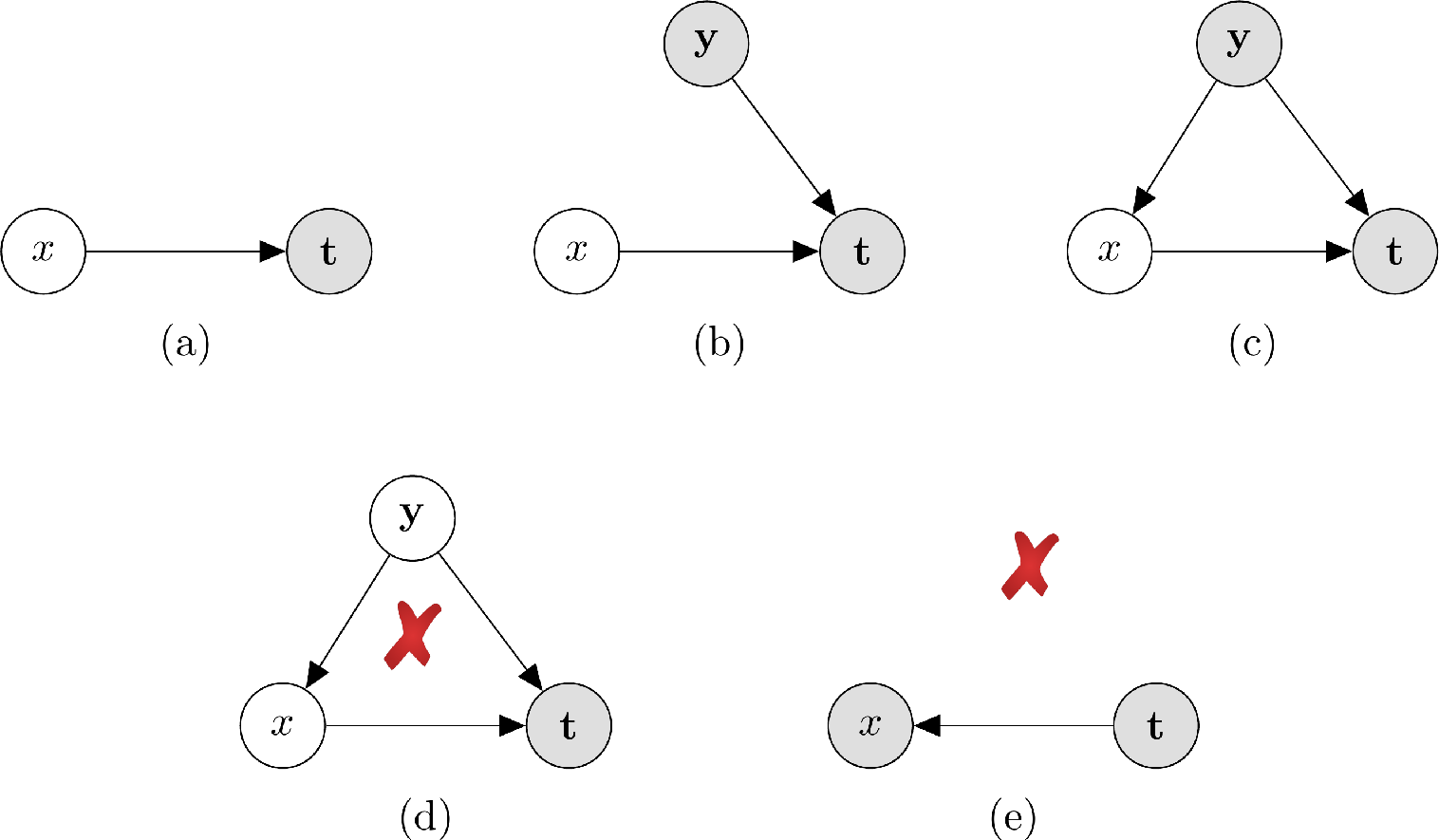}
  }
  \caption{
  \revision{
  The causal diagrams of the models considered in this paper
  are illustrated in the top row:
  (a) basic model encoding how the unknown variable of interest $x$ 
  generates
  an acquired brain scan $\fat{t}$; 
  (b) the model with additional known covariates $\fat{y}$ included;
  and 
  (c) the model where $\fat{y}$ are known confounders.
  The arrows indicate causal relationships, and variables in empty vs.~shaded circles are unknown vs.~observed, respectively.
  The bottom row illustrates two 
  cases that are not considered in this paper:
  (d)
  a model with confounders that are not observed;
  and
  (e) a decoding model where the direction of causality is reversed.
  }
  }
  \label{fig:graphs}
\end{figure}

} 

\revision{
\subsection{Scenarios not considered}

\noindent
Scenarios other than those shown in Fig.~\ref{fig:graphs}(a)-(c) are not considered in this paper. 
For instance, cases where confounders are not observed (Fig.~\ref{fig:graphs}(d)) would bias the estimated causal effect of $x$ on $\fat{t}$ to also include the effect of $\fat{y}$ on $\fat{t}$,
which would lead to misleading visual explanations (see Sec.~\ref{sec:covariates} for an example).
We also do not consider
\emph{decoding} models (discriminative models, illustrated in Fig.~\ref{fig:graphs}(e)) in which the direction of causality is reversed, i.e., where the brain changes 
visualized in $\fat{t}$ are the \emph{cause} (rather than the \emph{effect}) of $x$. 
%
In experimental neuroscience, 
such models are used to investigate the mapping from functional or structural anatomy to perceptual and behavioural consequences or their deficits~\citep{friston2008bayesian,polsterl2021estimation,chevalier2021decoding}.
%
In 
these 
applications,
the 
focus is on
elucidating the 
mapping between $\fat{t}$ and $x$
rather than on subject-level prediction,
as $x$ is already known.

} 

\revision{
\subsection{Related work}
} 
%

\noteToSelf{%
Relaxation of strong modeling assumptions in simple naive Bayes classifier papers [ Ng, Domingos ]. Have been shown to have strong prediction performance, especially [ but not only: Domingos also shows that wrong models can give correct decsion boundaries ] for moderate N. This is relevant because N moderate in many practical neuroimaging scenarios. [ Maybe mention here our bias-variance trade-off experiment -- Ng/Domingos don't show corresponding experiments, though ]
}
\noindent

%
%

\noindent
The method we propose can be viewed as a generalization of naive Bayesian classifiers,
in which the strong conditional independence assumption between input features is relaxed: 
When the number of latent variables is artificially clamped to zero in our method, 
the resulting predictor will devolve into a ``naive'' one.
Naive methods have previously been shown to 
have surprisingly strong
prediction performance in scenarios where the size of the training set is limited~\citep{domingos1997optimality,ng2002discriminative},
in part because their simple structure prevents overfitting~\citep{domingos1997optimality,domingos2012few}.
Our findings indicate that this property also holds for the proposed method: 
In training regimes with up to a few thousand subjects
\revision{-- the typical scenario in many applications --}
we obtain prediction accuracies that rival those of 
the best image-based prediction methods available to date.

\noteToSelf{
Previous attempts to combine prediction performance of discriminative methods with interpretability of generative models: DavatzikosMICCAI and Haufe. DavatzikosMICCAI forces discriminative and generative weight maps to be identical, wheras we have a interpretable generative/causal model that can easily be inverted to get good predictions without having to constrain/contaminate/bias generative/causal map. Haufe has recipe for turning discriminative weights [in linear models] into generative ones -- can be obtained as a special case of our method by training on predicted rather than true labels [ ref to equation ] -- however if true labels are available generative maps become independent of the discriminative weights.
}
\noindent

In its most basic (linear) form, the proposed method generates two 
spatial maps: 
a generative one that is suitable for human interpretation,
and 
a discriminative one, 
computed from the generative one, that the method uses to make predictions.
%
The distinct role 
in 
interpretation vs.~prediction
of these two different type of maps 
has been recognized 
before in the literature.
%
In~\citep{haufe2014interpretation},
for instance,
the authors proposed a technique for computing a linear generative map that is compatible with a 
discriminative one
\revision{and that is advocated to be more interpretable.
However, as we demonstrate in~\ref{sec:appendix_haufe},
this technique can be highly misleading 
as its always generates the same result irrespective of which image areas are actually used in the prediction computations, as long as the predictions themselves are accurate
(see Fig.~\ref{fig:haufe}).
}
%
%
%
At the other end of the spectrum,
in~\citep{varol2018generative} 
the authors 
developed a method in which the discriminative and the generative maps are forced to be identical.
Although good 
performance was reported,
the requirement to also make good predictions
will inevitably bias the generative map, 
potentially limiting its 
validity
as a tool for 
meaningful
neuroanatomical
interpretation.

%
%
%
  
\noteToSelf{%
Adrian Dalca, Daniel Alexander, Glocker NeuroIPS, [ Prince MICCAI? ], \ldots simulate subject-level aging process [careful: Dalca is only age-specific templates I think]. If causal model [ Glocker NeuroIPS, [ Prince MICCAI? ] ] counter-factuals, as in our method. However, not [demonstrated to be] inverted to yield accurate subject-level predictions.
}

\noteToSelf{
Other generative methods that are inverted: Wilms and Kilian's VAE for age. Like us they use latent variable models, but ``decoded'' nonlinearly [ in a deep learning setting ] [instead of linearly as in ours ].
[ In the following, not sure yet in how much depth we want to criticize these methods. We could also just say that this requires approximations in training and inverting (VAE), and in the age-specific template generation and of the input data itself (Wilms) -- contrasting this with closed-form equations of the linear model we propose for generating templates/group models, counterfactuals and predictions ]
Wils is flow-based and therefore dimensionality of input space is restricted to dimensionality of latent space, requiring PCA-based preprocessing of the input data which makes [ can make ] visualizing causal effect of age on brain shape more difficult to interpret [ overly smooth [ cf. their fig 2 vs.~ our generative maps ] ].
[ Also: target variable is subjected to nonlinear mapping, so age-specific template generation requires Monte Carlo sampling ]
VAE can be seen as a [direct] nonlinear version of the method proposed here [ show equation somewhere in a footnote, perhaps when we mention it as benchmark, and refer to it here ], which requires [ variational ] approximations to be made for training and inverting the model [, which is not needed ].
[ (Experimental comparisons of our method with the VAE are shown in section XYZ ) ]
[ Probably not worth including: Because the target variable is subjected to nonlinear mapping in both methods, visualizing the causal group effect is cumbersome ] 
  
[As opposed to our method, where ] This nonlinearity makes approximations [variational and Monte Carlo] necessary during training . Furthermore,  because the target variable 
VAE can be seen as a nonlinear version 
  [ (Experimental comparisons of our method with the VAE are shown in section XYZ ) ]
}
\noindent

Several methods exist that, like the proposed method, 
allow
one to generate synthetic images simulating the effect of specific variables of interest on brain shape 
--
either on a population level (e.g., age-specific brain templates~\citep{dalca2019learning,pinaya2022brain,wilms2022invertible,zhao2019variational}) 
or for individual subjects (e.g., artificially aging an individual brain~\citep{pawlowski2020deep,ravi2019degenerative,wilms2022invertible,xia2021learning}).
However, only a few of these methods are 
designed to 
also
be
``inverted'' to provide accurate single-subject predictions.
%
%
Like our method, 
both~\citep{zhao2019variational} and~\citep{wilms2022invertible} use latent variable models, but, unlike ours, they ``decode'' 
these latent variables 
using neural networks:
\citep{zhao2019variational} is based on a variational autoencoder (VAE),
whereas 
\citep{wilms2022invertible}
uses normalizing flows.
The resulting nonlinearities 
increase the expressiveness of the models,
but 
come at the price of 
additional computational complexity and the need for 
various approximations
during training and inference~\citep{zhao2019variational} 
or even of the input data itself%
~\citep{wilms2022invertible}.
In contrast,
predicting, generating conditional templates,
computing
counterfactuals,
and even training involve only evaluating analytical expressions 
with the proposed method.
%
An experimental comparison of our method with the VAE of~\citep{zhao2019variational}, detailed in Sec.~\ref{sec:prediction_performances},
suggests that this simplicity does not 
come with
a 
loss in
prediction accuracy.



\bigskip
\revision{
\section{Basic Linear Version}
\label{sec:method}
}

\noindent
In this section, we describe 
\revision{%
the proposed method in its most elementary form: 
a simple causal relationship between a 
variable 
of interest
and the resulting image 
-- 
the situation depicted in
Fig.~\ref{fig:graphs}(a)
--
that is furthermore assumed to be linear.%
}
%
\revision{
More complex models, with nonlinear dependencies on the variable of interest or the inclusion of subject-specific covariates and confounders,
will be presented in Sec.~\ref{sec:extensions}. 
}

\bigskip
\subsection{Generative model}
\label{sec:generativeModel}

\begin{figure*}
  \setlength{\myWidth}{\linewidth}
        \includegraphics[width=\myWidth]{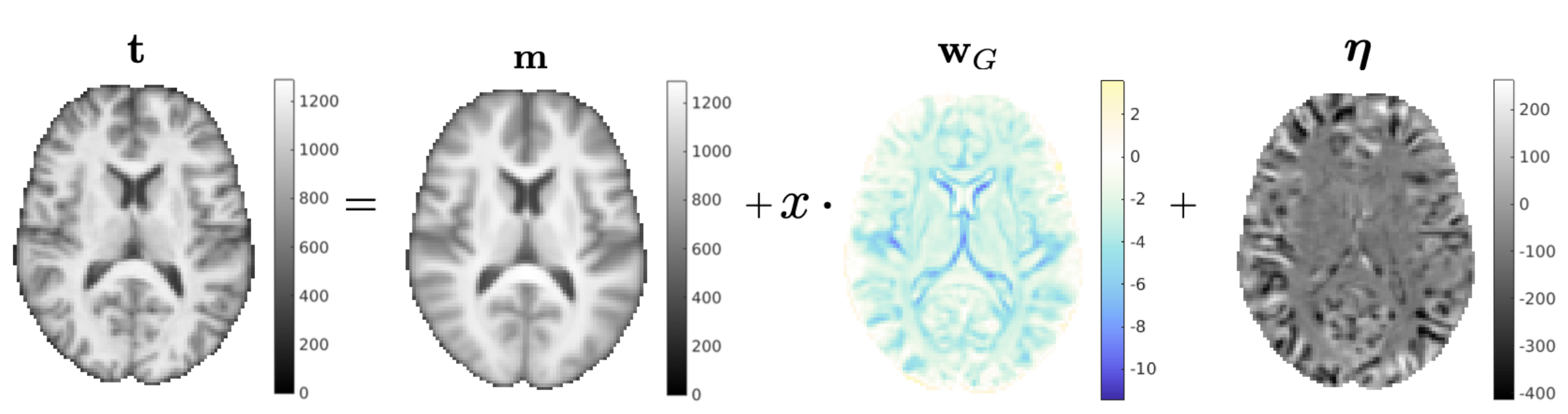}
  \caption{Example of the forward model~\eqref{eq:decomposition}, 
  applied to 
  modeling 
  the effect of age on brain morphology.
  Here $x$ denotes the difference between the age of the subject and the average age in a training set.
  }
  \label{fig:generativeModelWithRealImages}
\end{figure*}

\noindent
Let $\fat{t} \in \mathbb{R}^J$ denote a 
a vector that contains the intensities in the $J$ voxels of a subject's image,
and 
$x$ 
a
scalar variable of interest
about that subject 
(such as their age or gender).
A simple generative model,
illustrated in Fig.
~\ref{fig:generativeModelWithRealImages},
is then of the form
\begin{equation}
  \fat{t}=\fat{m} + x \fat{w}_G + \bldgr{\eta}
  .
  \label{eq:decomposition}
\end{equation}
Here 
$\fat{w}_G \in \mathbb{R}^J$
is a spatial weight map
-- referred to as the \emph{generative} weight map in the remainder --
that reflects 
how strongly the variable of interest $x$ is expressed in the voxels of $\fat{t}$:
Assuming a causal relationship between $x$ and $\fat{t}$, 
it encodes how a unit increase $x$ changes each voxel's intensity, on average.
Further,
$\fat{m} \in \mathbb{R}^J$
is a spatial template of intensities at baseline (i.e., when $x=0$),
and 
$\bldgr{\eta} \in \mathbb{R}^J$ is a random noise vector, assumed to be 
Gaussian distributed with zero mean and covariance $\fat{C}$.
For notational convenience
we will 
collect the two spatial weight maps $\fat{m}$ and $\fat{w}_G$ in a single 
matrix $\fat{W} = ( \fat{m}, \fat{w}_G )$ for the remainder of the paper.

%
Note that this is the model commonly assumed in traditional mass-univariate brain mapping techniques, 
such as voxel- and deformation-based morphometry \citep{ashburner2000voxel,chung2001unified}, where diagonal $\fat{C}$ is assumed and $\fat{w}_G$ is analyzed with statistical tests to
reveal brain regions with significant effects.
%
%
%
In contrast, here we assume that $\fat{C}$ has spatial structure, allowing us, besides interpreting $\fat{w}_G$, to accurately predict $x$ from $\fat{t}$ by inverting the model, as shown below.


\subsection{Making predictions}
\label{sec:makingPredictions}
\noindent
When 
the parameters of the model ($\fat{W}$ and $\fat{C}$) are known,
the 
unknown
target 
variable 
$x^*$ of a subject with image $\fat{t}^*$ 
can be inferred by inverting the model using Bayes' rule.
For a binary target variable $x^* \in \{0,1\}$
\revision{with prior probability $p(x^*)$,}
it is well-known that 
the target posterior distribution takes the form of a logistic regression classifier~\citep{hart2000pattern}
%
%
%
\revision{as shown in \ref{sec:appendix_predictions}:}
\begin{equation}
  p( x^* = 1 | \fat{t}^*, 
  \fat{W}, 
  \fat{C} )
  = \sigma\big( \fat{w}_D^T \fat{t}^* + w_o \big)
  ,
  \label{eq:posterior_classification}
\end{equation}
%
%
%
where
\begin{equation}
  \fat{w}_D = \fat{C}^{-1} \fat{w}_G
  \label{eq:discriminativeWeights}
\end{equation}
are a set \emph{discriminative} spatial weights,
$\sigma( a) = 1/(1+e^{-a})$ 
denotes the logistic function,
and
$
w_o = -\fat{w}_D^T( \fat{m} + \fat{w}_G/2 )
$
$
\revision{
+ \log \left[ p(x^*\!\!=\!\!1) / p(x^*\!\!=\!\!0) \right]
}
$.
%
The 
maximum a posteriori (MAP)
estimate
of $x^*$ is therefore $1$ if 
\begin{equation}\label{eq:prediction_classification}
  \fat{w}_D^T \fat{t}^* + w_o > 0,
\end{equation}
and $0$ otherwise.
\revision{
In the remainder of the paper, we will assume 
equal priors: $p(x^*\!\!=\!\!1) = p( x^*\!\!=\!\!0) = 0.5$ unless stated otherwise.
}

\begin{figure*}[t!]
\centering
  \setlength{\myWidth}{\linewidth}
      \includegraphics[width=0.52\linewidth]{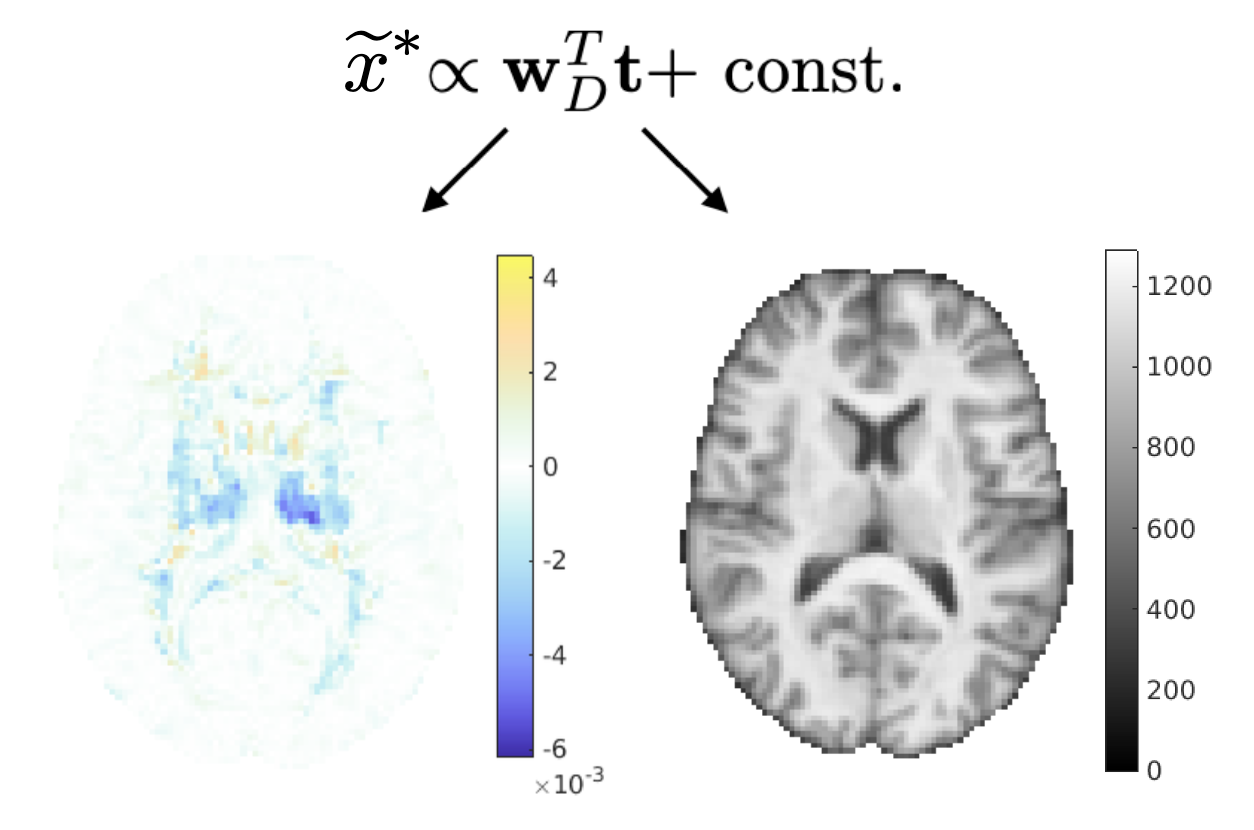}
  \caption{Illustration of how a subject's age is estimated by inverting the model shown in Fig.~\ref{fig:generativeModelWithRealImages}.
  }
  \label{fig:modelInversionWithImages}
\end{figure*}

For a continuous target variable with 
a flat prior
$p(x^* ) \propto 1$, 
the 
posterior distribution is Gaussian with 
variance 
\begin{equation}
  v =
  \left( \fat{w}_G^T \fat{C} ^{-1}  \fat{w}_G\right)^{-1}
  \label{eq:variance}
\end{equation}
and
mean
\begin{equation}
  \tilde{x}^*
  =
  v
  (
  \fat{w}_D^T\fat{t}^* + b_0
  ),  
  \label{eq:mean}
\end{equation}
where
$
  b_0 = -\fat{w}_D^T \fat{m}
$
(cf.
~\ref{sec:appendix_predictions}).
The predicted value of $x^{*}$ is therefore given by (\ref{eq:mean}),
which again involves taking the inner product of the discriminative weights $\fat{w}_D$ with $\fat{t}^{*}$.
An example of model inversion in case of age prediction is shown in Fig.~\ref{fig:modelInversionWithImages}. 

\subsection{Model training}

\noindent
In practice the model parameters 
$\fat{W}$ and $\fat{C}$ 
need to be estimated from training data.
Given $N$ training pairs $\{\fat{t}_n, x_n\}_{n=1}^N$,
their maximum likelihood (ML) 
estimate 
is 
obtained by maximizing
the 
marginal likelihood
\begin{equation}
  p\left( \{ \fat{t}_n \}_{n=1}^N | \{ x_n \}_{n=1}^N, \fat{W}, \fat{C} \right) 
  =
  \prod_{n=1}^{N} \mathcal{N}\left(\fat{t}_n|~ \fat{W} \bldgr{\phi}_n ,\fat{C}\right)
  \label{eq:marginalLikelihood}
\end{equation}
with respect to 
$\fat{W}$ and $\fat{C}$, where we have defined $\bldgr{\phi}_n = (1, x_n )^T$.
For the spatial maps,
the 
solution
is given in closed form (see
\ref{sec:appendix_W}):
\begin{equation}
  \fat{W}
  =  \left(\sum_{n=1}^{N} \fat{t}_n \bldgr{\phi}_n^T\right) \left( \sum_{n=1}^{N} \bldgr{\phi}_n \bldgr{\phi}_n^T \right)^{-1}
  .  
  \label{eq:W}
\end{equation}
%
%
However, obtaining the noise covariance matrix $\fat{C}$ directly by ML estimation is problematic: 
$\fat{C}$ has $J(J+1)/2$ free parameters, 
which is
orders of magnitude more than there are training samples
(recall that $J$ is the number of voxels).
%
%
To be able to control the number of parameters while still capturing the dominant correlations in the noise,
we impose
%
%
a specific structure on $\fat{C}$ by using a latent variable model known as factor analysis
\citep{bishop2006pattern}.
%
%
%
%
In particular,
we model the noise as
\begin{equation}\label{eq:eta}
  \bldgr{\eta}=\fat{V} \fat{z} + \bldgr{\epsilon},
\end{equation}
where $\fat{z}$ is a small set of $K$ unknown latent variables distributed as
$
p(\fat{z})=\mathcal{N}(\fat{z}|\fat{0},\mathbb{I}_K)
$,
$\fat{V}$ contains $K$ corresponding, unknown spatial weight maps,
and $\bldgr{\epsilon}$ is 
a zero-mean Gaussian distributed error with unknown diagonal covariance $\bldgr{\Delta}$.
Marginalizing over $\fat{z}$ 
yields
a zero-mean Gaussian noise model~\citep{bishop2006pattern} 
with 
covariance matrix
$$
  \fat{C}=\fat{V} \fat{V}^{T}+\bldgr{\Delta},
$$
which 
is now controlled by a reduced set of parameters $\fat{V}$ and $\bldgr{\Delta}$.
The number of 
columns in $\fat{V}$
(i.e., the number of latent variables $K$) is a hyperparameter in the model that needs to be tuned experimentally.

%
%

Plugging in the ML estimate of 
$\fat{W}$ 
given by (\ref{eq:W}),
the parameters $\fat{V}$ and $\bldgr{\Delta}$ maximizing the marginal likelihood (\ref{eq:marginalLikelihood}) 
can be estimated using an Expectation-Maximization (EM) algorithm~\citep{rubin1982algorithms}.
%
Defining 
\begin{equation}
  \bldgr{\eta}_n = \fat{t}_n - \fat{W} \bldgr{\phi}_n
  \label{eq:noiseVectorEstimation}
\end{equation}
as the noise vector of training subject $n$,
this yields an iterative algorithm that repeatedly evaluates the posterior distribution over the latent variables:
\begin{equation}\label{eq:posterior_zn}
p(\fat{z}_n|
\bldgr{\eta}_n, 
\fat{V}, \bldgr{\Delta})=\mathcal{N}(\fat{z}_n|\bldgr{\mu}_n,\bldgr{\Sigma}),
\quad \forall n
\end{equation}
where
$
\bldgr{\mu}_n=\bldgr{\Sigma} \fat{V}^T \bldgr{\Delta}^{-1} 
\bldgr{\eta}_n$
and 
$
\bldgr{\Sigma}=(\mathbb{I}_K+\fat{V}^T \bldgr{\Delta}^{-1} \fat{V})^{-1}
$,
and subsequently updates the 
parameters
accordingly:
\begin{equation}        \label{eq:updateV}
\fat{V} \gets
\left(\sum_{n=1}^{N}
\bldgr{\eta}_n 
\bldgr{\mu}_n^T \right) \left( \sum_{n=1}^{N}   \left( \bldgr{\mu}_n \bldgr{\mu}_n^T+\bldgr{\Sigma}\right)\right)^{-1}
\end{equation}
\begin{equation}
\bldgr{\Delta} \gets \mathrm{diag}
\left(
\frac{1}{N}
\sum_{n=1}^{N}
  \bldgr{\eta}_n
  \bldgr{\eta}_n^T
  -
  \fat{V}\frac{1}{N}\sum_{n=1}^{N} \bldgr{\mu}_n 
  \bldgr{\eta}_n^T 
\right)
.
  \label{eq:updateDelta}
\end{equation}
Here $\mathrm{diag}(\cdot)$ sets all the non-diagonal entries to zero.

\subsection{Practical implementation}

\noindent
The method outlined above
involves manipulating matrices of size $J \! \times \! J$.
Despite the high dimensionality (recall that $J$ is the number of voxels), 
computations can be 
performed efficiently
by
exploiting the structure of these matrices:
As detailed in 
~\ref{sec:appendix_implementation},
both training and predicting 
can be 
implemented in a way that only involves
the posterior covariance of the latent variables $\bldgr{\Sigma}$, which is of much smaller size $K \! \times \! K$.

In our implementation, we 
center
the target variable $x$, i.e., we 
use values from which 
the sample mean in the training set
has been subtracted.
As shown in~\ref{sec:appendix_haufe},
this has the 
advantage
that the estimated template $\fat{m}$ then
represents the anatomy of the ``average'' subject in the training set,
i.e., $\fat{m} = 1/N \sum_{n=1}^N \fat{t}_n$.
%
%
For estimating the parameters $\bldgr{V}$ and $\bldgr{\Delta}$ of the noise model, we 
first
perform a voxel-wise rescaling 
of 
the 
noise vectors
$\{\bldgr{\eta}_n\}_{n=1}^N$, 
such that each voxel has unit variance across the training subjects. 
We then initialize the EM algorithm by 
using a matrix with standard Gaussian random entries
for $\fat{V}$,
and the identity matrix
for $\bldgr{\Delta}$. 
Convergence of the EM procedure is detected 
when the relative change in the log marginal likelihood drops below $10^{-5}$ between iterations.
The
elements in the estimated $\fat{V}$ and $\bldgr{\Delta}$ are then rescaled back to the original intensity space
to obtain the final parameters of the noise model. 

The code for the proposed model is available at \url{https://github.com/chiara-mauri/Interpretable-subject-level-prediction}, with both Matlab and Python implementations.

%



\section{Experiments on Age and Gender Prediction}
\label{sec:age_gender_prediciton}

\noindent
In this section, we 
present
experiments on the task of predicting a subject's age and gender from their brain MRI scan. Specifically, we 
compare
the behavior of the 
\revision{basic linear 
model described in Sec.~\ref{sec:method}%
}
with
that of three state-of-the-art benchmark methods, when the size of the training set is varied.


\subsection{Experimental set-up}

\noindent
The benchmarks we used
consist of
a nonlinear-discriminative (SFCN \citep{peng2021accurate}), 
a linear-discriminative (RVoxM \citep{rvoxm}), 
and a nonlinear-generative (VAE \citep{zhao2019variational}) method
for image-based prediction.
Together with the proposed method, which is 
linear-generative
in the basic form analyzed here,
these 
benchmarks
form a representative sample of the 
spectrum of 
methods 
available to date.

We trained each of these methods on randomly sampled subsets of 
26,127 
T1-weighted scans of healthy subjects 
drawn from the UK Biobank~\citep{alfaro2018image}.
Each of these subjects was between 44 and 82 years old, 
and was scanned on one of three identical 3T scanners using the same MRI protocol.
In our experiments, 
we used the skull-stripped and bias-field corrected T1-weighted scans 
that are 
made publicly available,
which are computed using a nonlinear registration with the MNI152 template
(see~\citep{alfaro2018image} for details).
We 
took advantage of these nonlinear registrations %
%
to warp 
and resample
all the scans 
to the 
same 
MNI152 template space
using linear interpolation. 
The resulting 
1mm isotropic 
T1-weighted images then formed the input of the 
various
image-prediction algorithms.

%
%
To analyze the behavior of the different methods 
when the training set size is gradually increased from $100$ to almost $10,000$ subjects,
we trained each method multiple times for each training set size.
Specifically, each 
training run 
was repeated 10 times, with different randomly sampled training subjects, except for the larger training sizes ($N\!>\!1,000$)
where the number of repetitions was limited to 3 to reduce the computational burden.
After training, 
the prediction performance of
each
model
was  
evaluated 
on a fixed set of $1,000$ randomly sampled test subjects not overlapping with the training subjects. 
\revision{%
The size of this test set (and of a separate validation set, see below) was chosen 
to be similar to 
the one
used in%
~\citep{peng2021accurate} 
so that our results could be compared to those reported in that paper.
}%
For the age prediction task,
the Mean Absolute Error (MAE)~\footnote{MAE is defined as  the absolute difference between predicted and real value of the target variable, averaged across all test subjects.}
was used as the evaluation criterion, 
whereas 
the average classification accuracy was used in the gender prediction experiments.

\revision{%
Consistent with the} set-up of~\citep{peng2021accurate},
we also had a separate, fixed validation set of $500$ randomly sampled subjects.
This was used
to tweak 
\revision{the hyperparameter(s)} 
for each method 
(see details below),
using a grid search to optimize MAE (for age) and classification accuracy (for gender)
for
each training run.
\revision{%
The details about the hyperparameter values considered in the grid search 
for each method 
are provided in 
Sec.~3 of the supplementary material.%
}

\subsection{Implementation of the benchmark methods}
\label{sec:benchmarks}

\begin{figure*}[p]
  \centering
  \includegraphics[trim={1cm 1cm 1cm 1cm},clip,width=0.72\linewidth]
{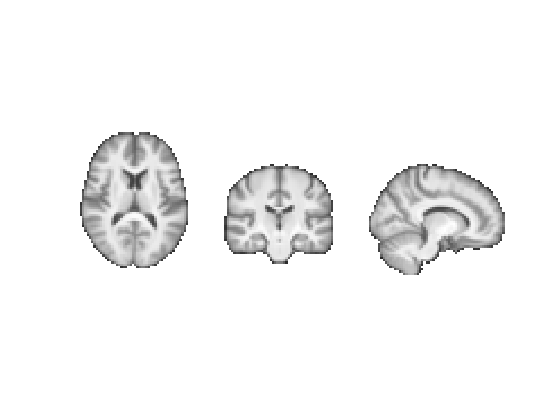}
  \vspace{-25.00mm}
  \caption{
  The estimated 
  template $\fat{m}$ (the average image in the training set)
  when the model is trained on $N=2,600$ subjects 
  in an age prediction task.
  }
  \label{fig:m}
\end{figure*}

\begin{figure*}[p]
  \centering
  \setlength{\myWidth}{0.30\linewidth}
  \begin{tabular}{cccc}
    \centering
    \hspace{-10.00mm}
    \includegraphics[trim={0cm 1cm 0cm 0cm},clip,width=0.25\linewidth]{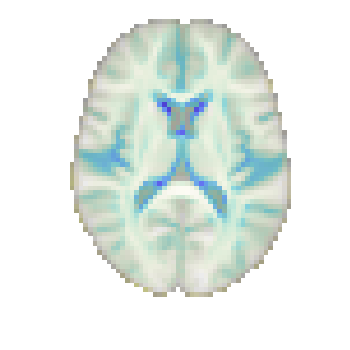} &
      \hspace{-8.00mm}
    \includegraphics[trim={0cm 1.5cm 0cm 0cm},clip,width=0.25\linewidth]{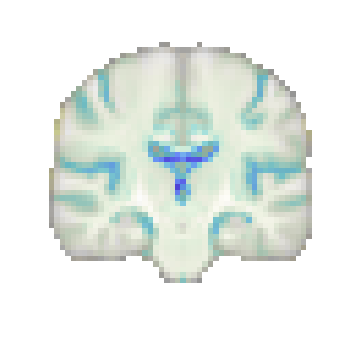} &
    \hspace{-3.00mm}
    \includegraphics[trim={0cm 0.5cm 0cm 0cm},clip,width=0.25\linewidth] {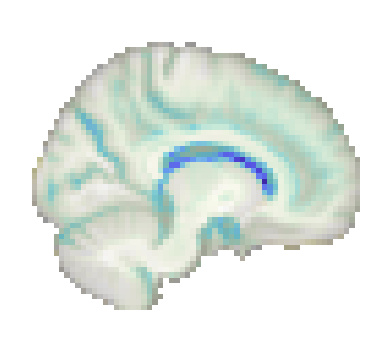} &
        \includegraphics[trim={0cm 0cm 0.3cm 0cm},clip,width=0.04\linewidth]{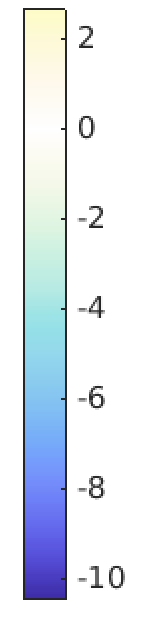} 
          \\
   \hspace{-10.00mm}
  \includegraphics[trim={0cm 1cm 0cm 0cm},clip,width=0.25\linewidth]{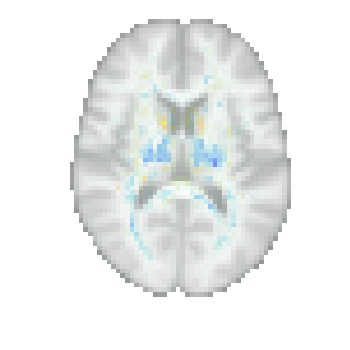} &
 \hspace{-8.00mm}
    \includegraphics[trim={0cm 1.5cm 0cm 0cm},clip,width=0.25\linewidth]{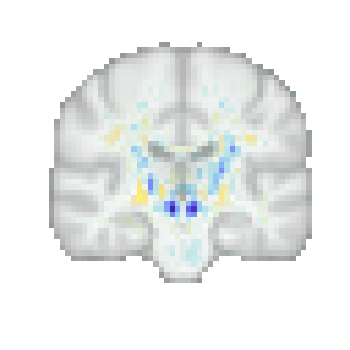} &
    \hspace{-3.00mm}
      \includegraphics[trim={0cm 0.5cm 0cm 0cm},clip,width=0.25\linewidth]{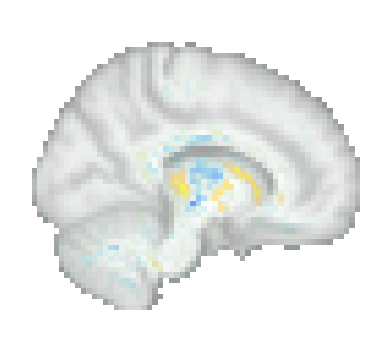} &
    \hspace{2.00mm}
     \includegraphics[trim={0.0cm 0.0cm 0.0cm 0cm},clip,width=0.05\linewidth]{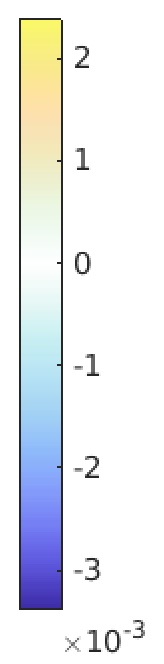} 
      \end{tabular}  
 \caption{
  Top: the generative map $\fat{w}_G$ -- expressing the effect of aging --
  estimated from $N=2,600$ subjects, 
  overlaid on the 
  template of Fig.~\ref{fig:m}.
  Voxels with zero weight are transparent.
  Bottom: the corresponding discriminative map $\fat{w}_D$ that is used for making age predictions. The discrepancy between these two maps is analyzed in Sec.~\ref{sec:interpretability}.
  }
  \label{fig:maps_proposed_method2}
\end{figure*}

\begin{figure*}[p]
  \centering
    \setlength{\myWidth}{\linewidth}
     \setlength{\fboxrule}{2pt}
     \fcolorbox{white}{white}{%
     \includegraphics[trim={0cm 0cm 0cm 0cm},clip,width=\myWidth]
{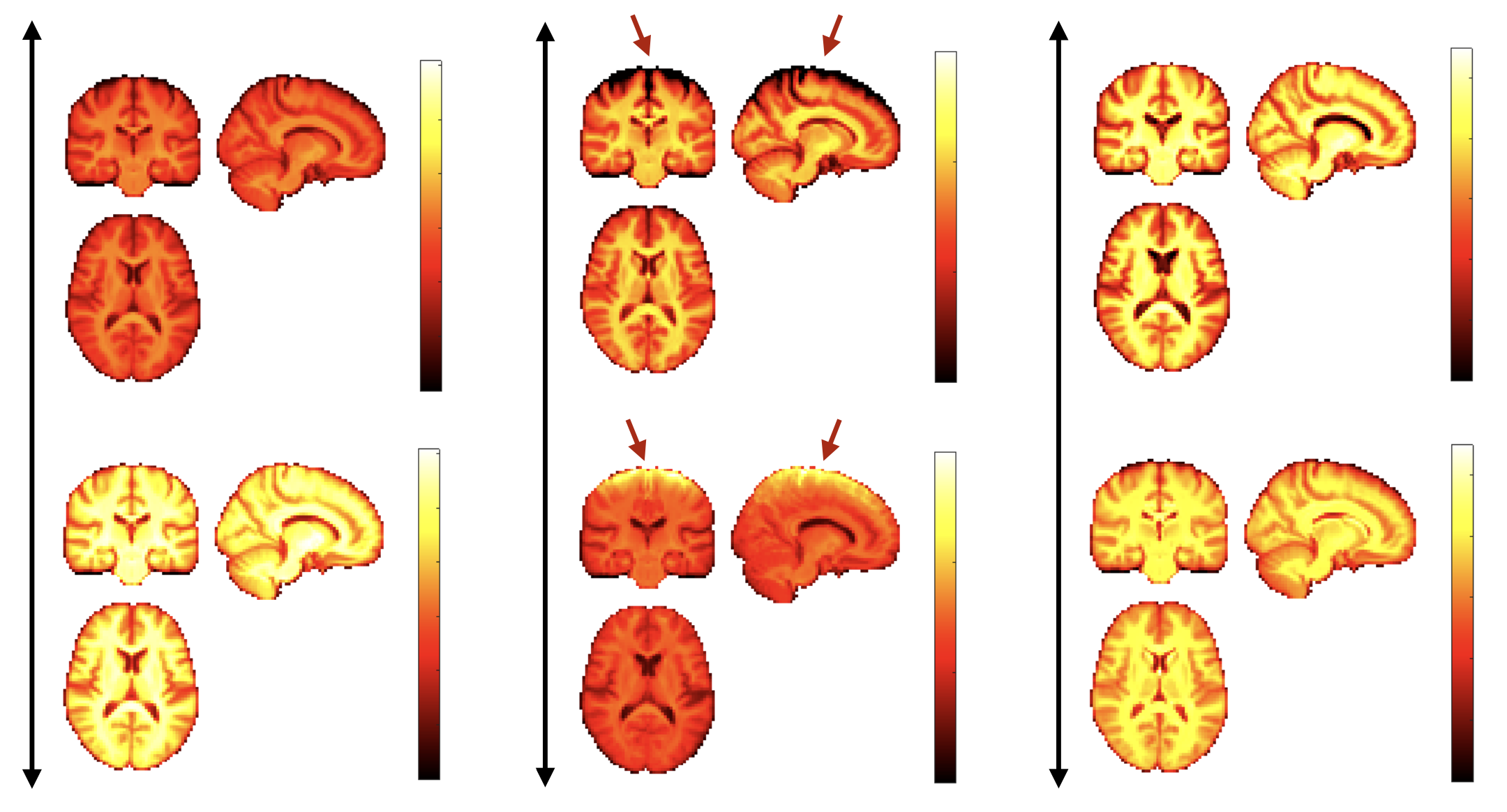} 
     }
     \\
    \begin{tabular}{ccc}
      \centering
       \hspace{-5.00mm}
      $1^{st}$ mode of variation &
     \hspace{26.00mm}
     $2^{nd}$ mode of variation & 
      \hspace{26.00mm}
     $3^{rd}$ mode of variation
        \end{tabular}  
  \caption{
  Illustration of the major modes of variation captured by the linear-Gaussian noise model, when trained on $N=2,600$ subjects 
  in an age prediction task.    
  Each mode illustrates the effect of applying one of the three first eigenvectors of $\fat{V}\fat{V}^T$ (the component of $\fat{C}$ that has spatial structure) 
  on the 
  template of Fig.~\ref{fig:m}.
  \revision{
  (We note that directly showing the learned basis vectors in the columns of $\fat{V}$ is not meaningful, since 
  the model is invariant to rotations in the latent space~\citep{bishop2006pattern}:
  using $\fat{\tilde{V}} = \fat{V}\fat{R}$ 
  for an arbitrary 
  orthogonal matrix $\fat{R}$ yields the same covariance matrix 
  $\fat{\tilde{V}}\fat{\tilde{V}}^T$ = $\fat{V}\fat{V}^T$.%
  )
  }%
  The top and bottom figures show the template modified in 
  the positive and negative
  direction of the eigenvectors, respectively.
  Note that the first eigenvector seems to encode a general brightening/darkening of the image intensities, while the second one models residual bias fields that were not removed 
  in the preprocessing of the UK Biobank data.  
  %
  }
  \label{fig:eigenvectors}
\end{figure*}

\noindent
The four methods under comparison were implemented as follows:
\begin{list}{-}{\leftmargin=1em \itemindent=0em}

  \item \textbf{Proposed method}: 
  We 
  performed all
  the experiments with the proposed method in Matlab, 
  running on a Linux CPU 
  machine
  (Intel Xeon E5-2660V3 10 Core CPU 2.60GHz, 128GB RAM).
  To speed up computations, the number of voxels was reduced by masking out the background 
  (by thresholding the average of all images in each training set),
  and by downsampling 
  the input data
  to a 3mm isotropic resolution.
  This downsampling
  was found not to affect the prediction performance in pilot experiments
  \revision{(see supplementary material)}.
  For each training run, performance on the validation set was used to set the number of latent variables $K$, which is the only hyperparameter of the method.
  %
  An example of a model trained 
  this way
  is shown in Figs.~\ref{fig:m}, \ref{fig:maps_proposed_method2} and \ref{fig:eigenvectors}.
%
%

\item \textbf{RVoxM}: 
This is a discriminative method that 
imposes
sparsity and spatial smoothness 
on 
its 
weight map as a form of regularization~\citep{rvoxm}.
\revision{%
As reported in the supplementary material,
it yields competitive prediction performances
compared to other commonly used 
linear-discriminative methods, 
and was therefore chosen to represent this family of models
(although other methods could also have been chosen).
%
}%
We
used the Matlab code that is publicly available\footnote{\url{https://sabuncu.engineering.cornell.edu/software-projects/relevance-voxel-machine-rvoxm-code-release/}}, with 
some
adaptations to make it more efficient 
(by parallelizing part of the training loop)
and, for gender prediction, 
more robust to
high-dimensional input.
%
%
The same 
background masking, downsampling procedure
and computer hardware was used as for the proposed method
in the experiments.
The method has a single hyperparameter to control the spatial smoothness of the
weight map that it computes, which was tuned on the validation set for each training run.
%
%

     

%

\item \textbf{SFCN}:
This is the lightweight convolutional neural network proposed by~\citep{peng2021accurate}, who, to the best of our knowledge, have reported the best performance for 
brain
age prediction to date.
Since code for training this method is not publicly available,     
we modified the 
implementation
of~\citep{mouches2022multimodal}%
\footnote{\url{https://github.com/pmouches/Multi-modal-biological-brain-age-prediction/blob/main/sfcn\_model.py}} 
to
match 
the description provided 
in~\citep{peng2021accurate}
as closely as possible.
In particular, 
we 
replicated
the 
same
data augmentation scheme, 
network architecture, L2 weight decay and
batch size,
%
deviating only in the last network 
layer 
as discretizing age into 40 bins (as described in~\citep{peng2021accurate}) did not 
benefit prediction accuracy in our experiments.
%
%
%
%
%
The method directly takes 
1mm isotropic images as input,
and was 
therefore
run on a high-end GPU with a 
sufficiently
large amount of memory 
(NVIDIA A100 SXM4 GPU with 40 GB of RAM) in our experiments.
%
For each training run,
the validation set was used to determine the optimal number of 
training 
epochs
of this method.
%
%
%
%

\item \textbf{VAE}: 
This
is a generative method for age prediction \citep{zhao2019variational} with publicly available training code\footnote{\url{https://github.com/QingyuZhao/VAE-for-Regression}}.
      %
      %
It can be regarded as a generalization of the proposed method, where the latent variables are expanded \emph{nonlinearly} through a deep neural network, which precludes exact model inversions and makes computations more involved.
%
Since the method is designed to work with
images that are downsampled to 2mm isotropic resolution and 
cropped around the ventricles~\citep{zhao2019variational},
the comparison with this method was performed separately from the other two benchmarks, with both the VAE and the proposed method running on the same cropped 2mm volumes. 
\revision{%
%
We tested only training sizes in a similar range (from 100 to 400 training subjects) as the one used in the original paper (196 subjects), running the VAE on a RTX 6000 GPU with 65G of RAM.
Although the method originally only contains two hyperparameters (dropout factor and L2 regularization), for a fair comparison we also varied the number of latent variables (originally hardcoded to 16) as an extra hyperpameter. 
As for the other methods, the values of the three resulting hyperpameters 
were tuned on the validation set for each training run.
}

%
%
%
%
      
\end{list}

\myclearpage
\subsection{Prediction performance and training times}
\label{sec:prediction_performances}

\begin{figure*}[p]
\centering
\begin{tabular}{cc}
\hspace{-4mm}
       \includegraphics[trim={1.2cm 0 1.0cm 0}, clip=true,width=0.5\linewidth]{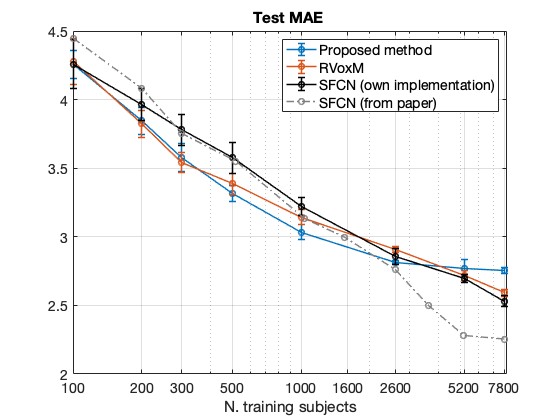}&
      \hspace{-5mm}
      \includegraphics[trim={1.0cm 0 1.0cm 0}, clip=true,width=0.5\linewidth]{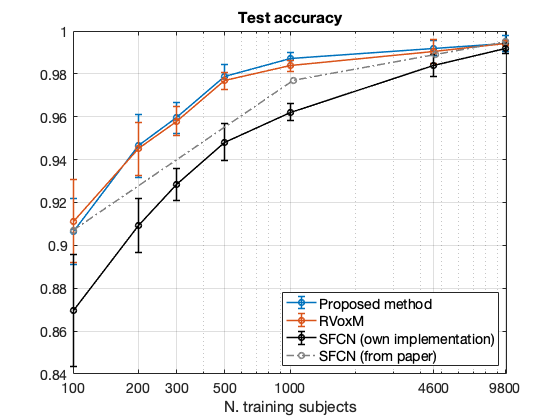}
\end{tabular}
\caption{
Prediction performance of the proposed method, RVoxM and SFCN for age (left) and gender (right) as the number of training subjects is varied.
For each method, the full line shows the average performance across multiple training runs, and the whiskers extend to one standard deviation away from the average.
The test MAE for age is indicated in years.
%
%
%
%
For reference, we also include the performance of SFCN as reported in~\citep{peng2021accurate},
although the results are not entirely comparable.
\todoDone{Mark says: Scaling/unit of y-axis = years?}
}
\label{fig:comparison_rvoxm_peng_new}
\end{figure*}

\noindent
Fig.~\ref{fig:comparison_rvoxm_peng_new} shows the prediction performance of the proposed method, RVoxM and SFCN as the number of training subjects 
is varied,
both for predicting age (Fig.~\ref{fig:comparison_rvoxm_peng_new} left) and gender (Fig.~\ref{fig:comparison_rvoxm_peng_new} right).
As expected, all prediction performances improve as the training size increases.
In the age prediction task, the proposed method yields generally the best performance when the training set size is 
up to 2,600 subjects,
after which SFCN is the better method.
For gender prediction, SFCN is clearly outperformed by both RVoxM and the proposed method.

For completeness, Fig.~\ref{fig:comparison_rvoxm_peng_new} also includes the results for SFCN as reported by its authors in~\citep{peng2021accurate}, where a similar experimental set-up 
as ours
was used.
It can be seen that, although our implementation closely followed the description provided by the authors, we were not always able to match their reported performance: Especially for gender prediction, and for age prediction with very large training sets, there are considerable discrepancies between the two implementations.
One possible explanation is that SFCN is a complex model with many more 
``knobs''
to be tuned correctly
than the proposed method, which has only a single hyperparameter.
Another explanation is that the experimental set-up is not entirely comparable: 
In~\citep{peng2021accurate}
the model is only trained 
on one training set
for each training size;
the subjects in their test set 
are
different from ours;
and they used affinely instead of nonlinearly registered scans 
(although the latter point is reported to yield only minimal differences in~\citep{peng2021accurate}, which we can confirm based on our own experiments).

Table~\ref{tab:trainingtimes} 
reports,
for the age prediction experiment, 
the training times required by 
the proposed method, RVoxM and SFCN.
SFCN is by far the slowest to train, requiring many hours even for very small training sizes, and several days for large ones. 
Up to training sizes of 
1,000 subjects,
the proposed method is considerably faster to train than RVoxM 
(dozens of minutes CPU time at most),
but slows down for larger training sizes.
This can be explained by the fact that the optimal number of latent variables in our model (hyperparameter $K$)
increases rapidly with the number of training subjects, as 
shown in Table~\ref{tab:hyperparameter}.
%
When 
interpreting
the training times of RVoxM and the proposed method on the one hand, and those of SFCN on the other, 
it should be taken into account that
the latter 
uses hardware acceleration
but also works
with
much larger (non-downsampled) 
input
volumes.

{
\renewcommand\arraystretch{1.4}
\addtolength{\tabcolsep}{+0.9mm}
\begin{table*}[p]
\begin{center}
\resizebox{1\textwidth}{!}{%
\begin{tabular}{|c|c|c|c|c|c|c|c|c|}
\hline
&  N=100 & N=200 & N=300 &N=500 &N=1,000 &N=2,600 & N=5,200 & N=7,800 \\ \hline
Proposed method  &1.20 min  &  0.67  min & 1.94 min  & 9.53 min  & 32 min & $\approx$3h& $\approx$15h &  $\approx$ 69 h\\ \hline 
RVoxM  &92 min &  66 min  & 75 min  & 76 min & 129 min & 127 min & $\approx$ 22 h & $\approx$ 21 h \\ \hline
SFCN & $\approx$ 8h & $\approx$ 11 h& $\approx$ 16 h & $\approx$ 18 h & $\approx$ 34h& $\approx$ 76h& $\approx$ 69h& $\approx$ 102 h\\ \hline
\end{tabular}
}
\caption{Training times 
in the age prediction task
for the proposed method, RVoxM and SFCN. 
These training 
times 
were evaluated
at the optimal value of a single hyperparameter for each method,
determined on an external validation set.
%
The reported times are the average across all training runs for each training set size $N$. 
}
\label{tab:trainingtimes}
\end{center}
\end{table*}
}

{
\renewcommand\arraystretch{1.3}
\addtolength{\tabcolsep}{+0.4mm}
\begin{table*}[p]
\tiny
\begin{center}
\resizebox{1\textwidth}{!}{%
\begin{tabular}{|c|c|c|c|c|c|c|c|c|}
\hline
&N=100 &  N=200& N=300 & N=500&N=1,000&N=2,600&N=5,200&N=7,800\\ \hline
$K$ 
&20 & 21 & 52 & 86 & 120 & 367 & 1,833 & 3,333 \\ \hline 
\end{tabular}
}
\caption{
Optimal number of latent variables $K$ of the proposed method when predicting age, as determined on an external validation set.
The reported number is the rounded average 
across all training runs for each training set size $N$.
}
\label{tab:hyperparameter}
\end{center}
\end{table*}
}

\begin{figure*}[p]
\setlength{\fboxrule}{2pt}
\centerline{\fcolorbox{white}{white}{
\includegraphics[trim={0cm 3cm 0cm 0cm},width=\columnwidth]{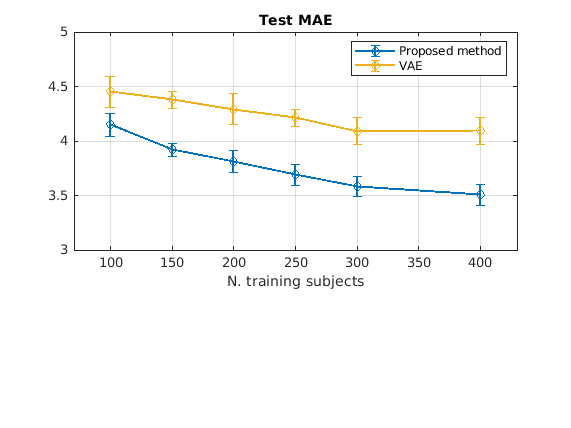}
}}
\caption{
Age prediction performance of the VAE and the proposed method, when applied to 2mm isotropic images cropped around the ventricles,
represented in the same way as in Fig.~\ref{fig:comparison_rvoxm_peng_new}.
}
\label{fig:mae_vae}
\end{figure*}

Finally, Fig.~\ref{fig:mae_vae} shows the age prediction performance of the VAE and the proposed method, as a function of the training set size.
The proposed method achieves better results for every 
training size
tested.
This suggests that, at least when 
only a few hundred subjects are available for training,
adding
nonlinearities in the 
method's
noise model
is not beneficial.
Furthermore, 
the VAE is 
considerably slower to train 
than the proposed method:
around 
\revision{8} 
min
GPU time for 200 training subjects,
\todoDone{still correct with new runs on different hardware?}
vs.~1 min 
CPU time with the proposed method.

\myclearpage
\subsection{Bias-variance trade-off}
\label{sec:biasVariance}

\noindent
More insight into the prediction performances reported in the previous section can be obtained using the so-called bias-variance decomposition.
Specifically for 
age,
which is a continuous variable,
a particular method's mean squared error (MSE) 
can be decomposed as follows~\citep{bishop2006pattern}:
\begin{multline}\label{eq:bias-var-E}
       \underbrace{\mathbb{E}_{D}\left[ \left(x^*-\tilde{x}^*_D \right)^2\right]}_{MSE}=
       \underbrace{ \left(x^*-\mathbb{E}_{D}\left[\tilde{x}^*_D\right] \right)^2}_{bias^2}+\\+\underbrace{\mathbb{E}_{D}\left[\left(\tilde{x}^*_D-\mathbb{E}_{D}\left[\tilde{x}^*_D\right]\right)^2\right]}_{variance}.
\end{multline}
Here 
$x^*$
denotes the real age of a given test subject, 
$\tilde{x}^*_D$ 
is 
the predicted age when the method is trained on a particular dataset $D$  
of a certain size,
and $\mathbb{E}_{D}[\cdot]$ denotes the average over multiple such training sets. 
%
In our set-up,
in which each method is trained multiple times using different randomly sampled 
subjects,
the \textit{bias} component 
in~\eqref{eq:bias-var-E}
reflects a systematic error 
that is persistent across 
the training 
runs, 
whereas
the 
\textit{variance} component
indicates 
how much the predictions change between the different training runs~ 
\citep{hart2000pattern,bishop2006pattern,domingos1997optimality}. 
Typically, 
flexible
models 
tend to have low bias but high variance, reflecting an \textit{overfitting} to the training data, while strongly constrained methods display the opposite behavior, resulting in \textit{underfitting} of the training data~\citep{hart2000pattern,bishop2006pattern,domingos1997optimality}.

\begin{figure*}[p]
\centering
\begin{tabular}{cc}
\includegraphics[trim={1.2cm 0 1.0cm 0}, clip=true,width=0.5\linewidth]{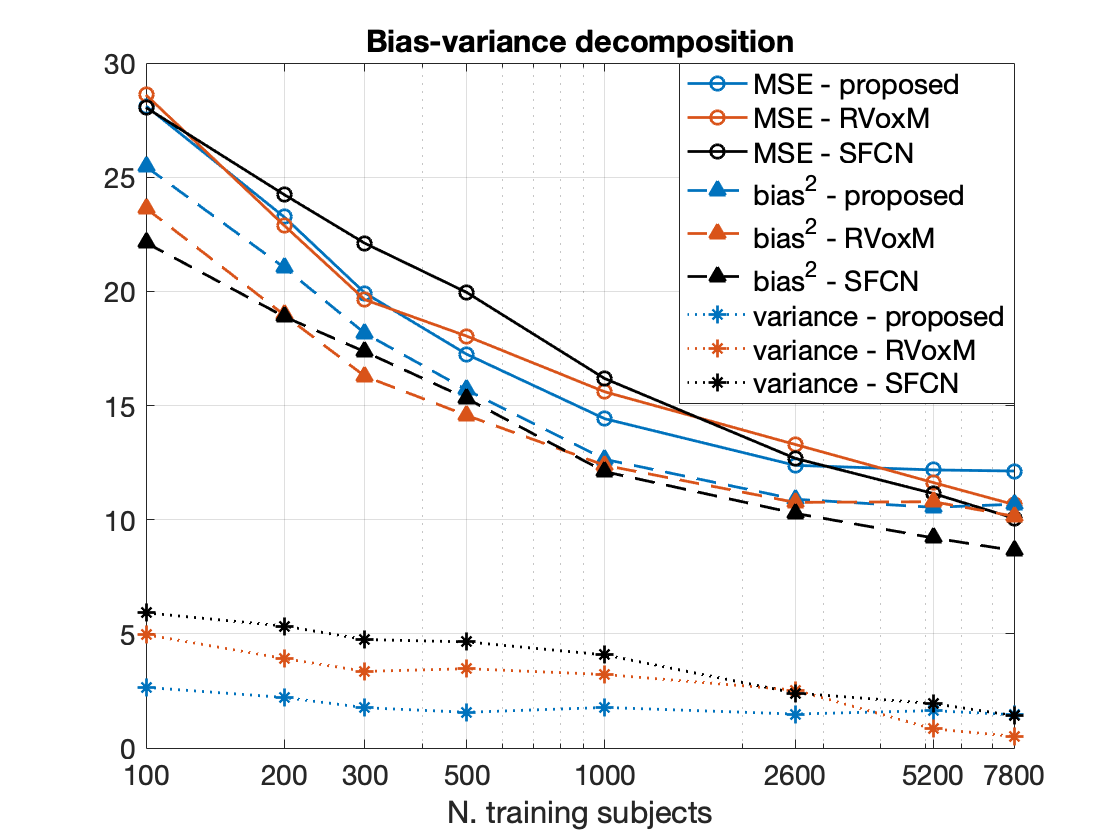} &
      {\setlength{\fboxrule}{2pt}\fcolorbox{white}{white}{\includegraphics[trim={1.0cm 0 1.0cm 0}, clip=true,width=0.46\linewidth]
      {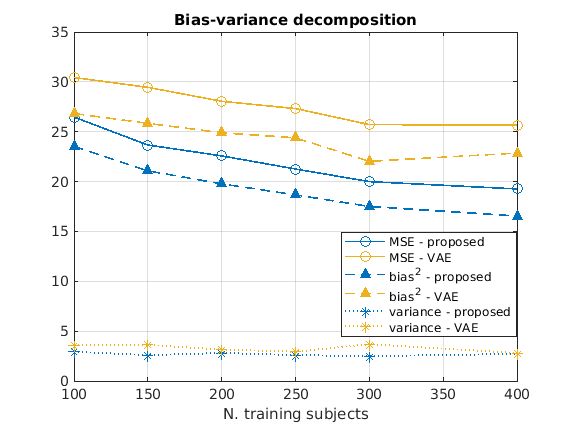}}}\\
\end{tabular}
\caption{Left: bias-variance decomposition 
for the age prediction results shown in 
Fig.~\ref{fig:comparison_rvoxm_peng_new} (left).
Right: the same for the results shown in Fig.~\ref{fig:mae_vae}.
}
\label{fig:bias_var}
\end{figure*}

Fig.~\ref{fig:bias_var} (left) shows how the bias, the variance and the resulting 
MSE 
change
in the age prediction experiment,
when the training set size is varied.
The 
various
curves 
were 
obtained
by averaging~\eqref{eq:bias-var-E}
across all test subjects
for our method 
(blue),
RVoXM (red) and 
SFCN (black).
%
The proposed
method generally has the highest bias among the three methods; 
however this is off-set by a lower variance, 
resulting in a
strong overall prediction performance
in training set sizes of up to 2,600 subjects.
%
%
Its
low variance is obtained by controlling the flexibility of the method:
when the number of training subjects is small, 
only a limited number of latent variables 
is selected
(see Table~\ref{tab:hyperparameter}), resulting 
in a simple, highly regularized model that successfully avoids overfitting
to the training data.
As the training size increases, the number of latent variables is allowed to grow, 
resulting in gradually more flexible models with decreased bias 
and therefore better prediction performance
(see Fig.~\ref{fig:bias_var} (left)).
However,
for very large training sets 
(over 
2,600 subjects), 
the method's strong modeling assumptions 
prevent it from decreasing its bias further,
and the 
nonlinear SFCN  can now take advantage of its flexibility
(lower bias)
without overfitting (variance comparable to the proposed method),
resulting in a better prediction performance.

%
%
%
%
%

Fig.~\ref{fig:bias_var} (right) shows the results of the bias-variance decomposition for the VAE and our method, when applied to age prediction from 2mm cropped data.
It can be seen that the VAE's lower prediction performance 
(reported in Sec.~\ref{sec:prediction_performances})
can be attributed to its significantly higher bias,
presumably
due to the variational approximations~\citep{kingma2013auto} that are used to invert its model.
\todoDone{update results, and make sure what we say here is still valid}

\myclearpage
\subsection{Interpretability analysis}\label{sec:interpretability}

\begin{figure*}
  \setlength{\myWidth}{0.21\linewidth}
 \centering
   \begin{tabular}{ccccc}
  \vspace{2mm}
  &
   $\fat{N=300}$ &
   $\fat{N=2,600}$ &
   $\fat{N=7,800}$ & 
   \\
  \vspace{2mm}
   Proposed ($\fat{w}_D$) &
    \includegraphics[align=c,trim={0cm 0cm 0cm 0cm},clip,width=\myWidth]{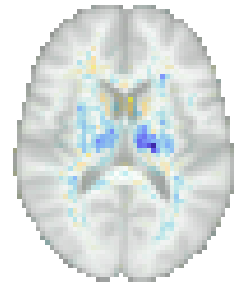} &
    \includegraphics[align=c,trim={0cm 0cm 0cm 0cm},clip,width=\myWidth]{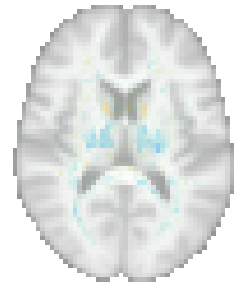} &
    \includegraphics[align=c,trim={0cm 0cm 0cm 0cm},clip,width=\myWidth]{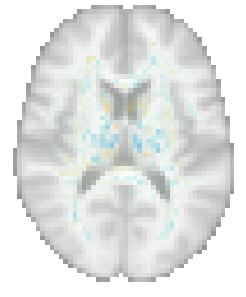} 
    & 
    \hspace{1mm}
     \includegraphics[align=c,trim={0cm 0cm 0cm 0cm},clip,width=0.066\linewidth]{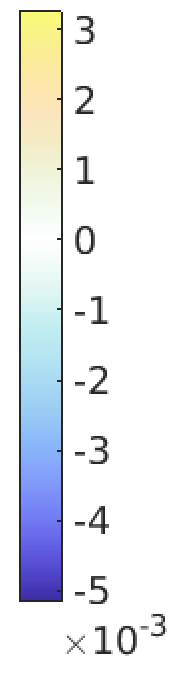}
   \\
   \vspace{2mm}
    RVoxM &
        {\includegraphics[align=c,trim={0cm 0cm 0cm 0cm},clip,width=\myWidth]{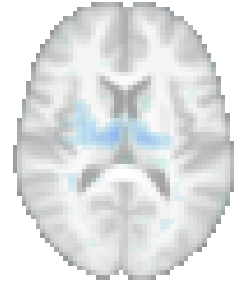}} &
    {\includegraphics[align=c,trim={0cm 0cm 0cm 0cm},clip,width=\myWidth]{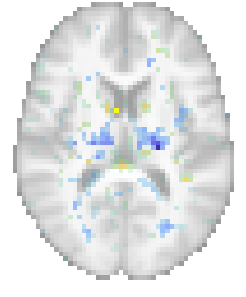}} &
    {\includegraphics[align=c,trim={0cm 0cm 0cm 0cm},clip,width=\myWidth]{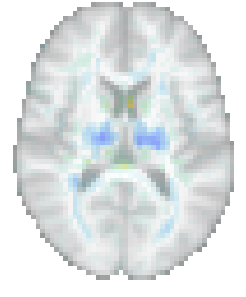}} &
    \hspace{1mm}
  {\includegraphics[align=c,trim={0cm 0cm 0cm 0cm},clip,width=0.06\linewidth]{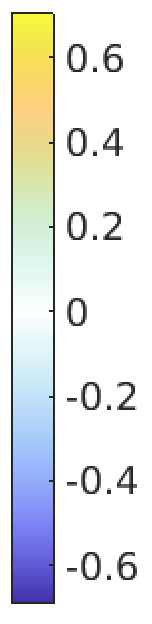}}
   \\
   \vspace{2mm}
   SFCN &
   \hspace{2mm}
    \includegraphics[align=c,trim={0cm 0cm 0cm 0cm},clip,width=0.22\linewidth]{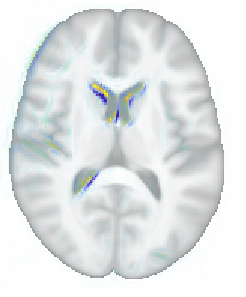} &
     \hspace{2mm}
    \includegraphics[align=c,trim={0cm 0cm 0cm 0cm},clip,width=0.22\linewidth]{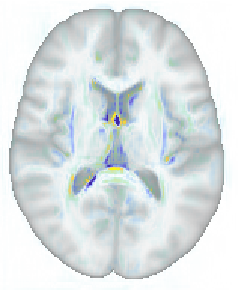} &
     \hspace{2mm}
    \includegraphics[align=c,trim={0cm 0cm 0cm 0cm},clip,width=0.22\linewidth]{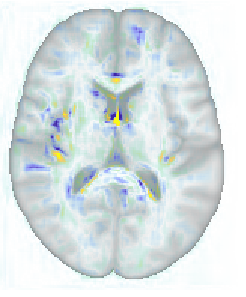} &
     \hspace{2mm}
     \includegraphics[align=c,trim={0cm 0cm 0cm 0cm},clip,width=0.075\linewidth]{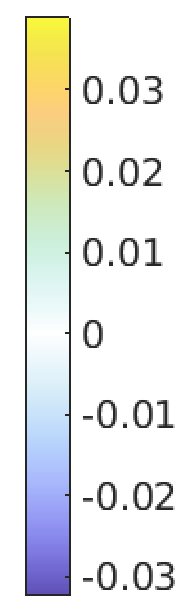}
   \\
  \end{tabular}  
  \caption{Discriminative maps of proposed method, 
  RVoxM, and SFCN (with SmoothGrad) obtained for age prediction with varying training set sizes, 
  overlaid on a template.
  Voxels with zero weight are transparent. 
  Since SmoothGrad yields subject-specific maps, we averaged across all test subjects to obtain population-level maps for SFCN.
  }
  \label{fig:selected_sets_combined}
\end{figure*}

\noindent
A key advantage of the proposed method over discriminative methods such as RVoxM and SFCN is that, in addition to the discriminative map $\fat{w}_D$ that it uses to make predictions, it also computes a generative map $\fat{w}_G$ 
that expresses the causal effect of the variable of interest on brain morphology.
To illustrate why this is important, 
Fig.~\ref{fig:selected_sets_combined} shows,
for three different training set sizes,
the discriminative map $\fat{w}_D$ computed by our method for predicting age, 
along with the corresponding discriminative map of RVoXM and  
 the SmoothGrad saliency map~\citep{smilkov2017smoothgrad}
-- a generalization of linear spatial maps to nonlinear methods~\citep{adebayo2018sanity} -- of SFCN.
%
The inconsistencies of these maps across both the training set sizes and the different methods,
and their overall lack of correspondence with the known 
neurobiology
of aging,
illustrate the 
difficulty 
of using discriminative maps for 
human interpretation.

\begin{figure*}[p]
  \setlength{\myWidth}{0.215\linewidth}
  \centering
  \begin{tabular}{cccc}
 \vspace{2mm}
   $\fat{N=300}$ &
   $\fat{N=2,600}$ &
   $\fat{N=7,800}$ & 
   \\
    \includegraphics[align=c,
    clip,width=\myWidth]{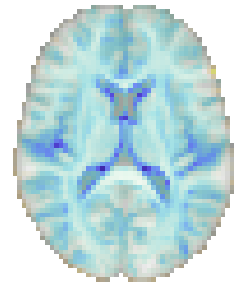} &
    \includegraphics[align=c,trim={0cm 0cm 0cm 0cm},clip,width=\myWidth]{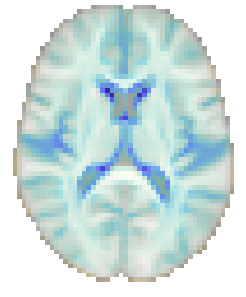} &
    \includegraphics[align=c,trim={0cm 0cm 0cm 0cm},clip,width=\myWidth]{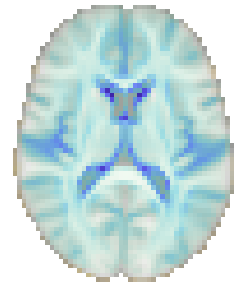} 
    & 
      \includegraphics[align=c,trim={0cm 0cm 0cm 0cm},clip,width=0.05\linewidth]{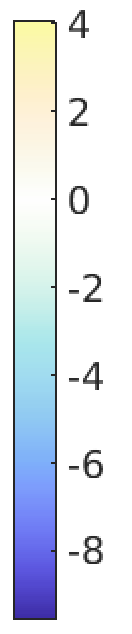}
    \\
  \end{tabular}  
  \caption{Generative maps $\fat{w}_G$ computed by the proposed method for age prediction, for different training set sizes.
  In addition to expressing known aging patterns in the brain, these maps also show consistency across the different training sizes.
%
}
  \label{fig:wGs}
\end{figure*}

More insight can be gained by examining the proposed method specifically,
since 
it uses
discriminative maps 
that
are derived from 
generative 
ones.
%
It is worth noting that 
estimating 
the 
generative maps
from training data 
is itself 
quite stable, since it 
merely
amounts to fitting two 
basis functions to 
hundreds of measurements in each voxel 
(see \eqref{eq:W}). 
Furthermore,
as illustrated in Fig.~\ref{fig:wGs},
the resulting 
maps are
intuitive to interpret,
since 
they show
typical age-related effects such as 
cortical thinning
and
ventricle enlargement~\citep{Fjell2009Cortex,fjell2010structural}.
%
When 
the discriminative maps are subsequently computed as 
$\fat{w}_D = \fat{C}^{-1} \fat{w}_G$,
however,
a strong dependency on the training set size is introduced, 
because the method explicitly 
controls the complexity of 
its noise model 
$\fat{C}$
in response to the size of the available training set
(the bias-variance trade-off of Sec.~\ref{sec:biasVariance}).
$\fat{C}$
can 
also
capture
peculiarities
in the data 
that may be relevant for improving prediction performance,
but not for human interpretation.
An example of this was shown in Fig.~\ref{fig:eigenvectors},
where 
overall brightness variations 
and residual MR bias field artifacts were picked up by the noise model.
Through $\fat{C}$, 
such 
noise
patterns 
can
find their way 
into $\fat{w}_D$,
producing hard-to-interpret spatial maps 
that 
no longer 
reflect 
the expected
age-related
brain
atrophy
patterns.
This is clearly 
illustrated in Fig.~\ref{fig:maps_proposed_method2},
where
the discriminative map $\fat{w}_D$
is contrasted with the 
corresponding
generative map $\fat{w}_G$.

\begin{figure*}[p]
  \centering
   %
 %
 %
 %
  \hspace{-0mm}
\includegraphics[width=\linewidth]{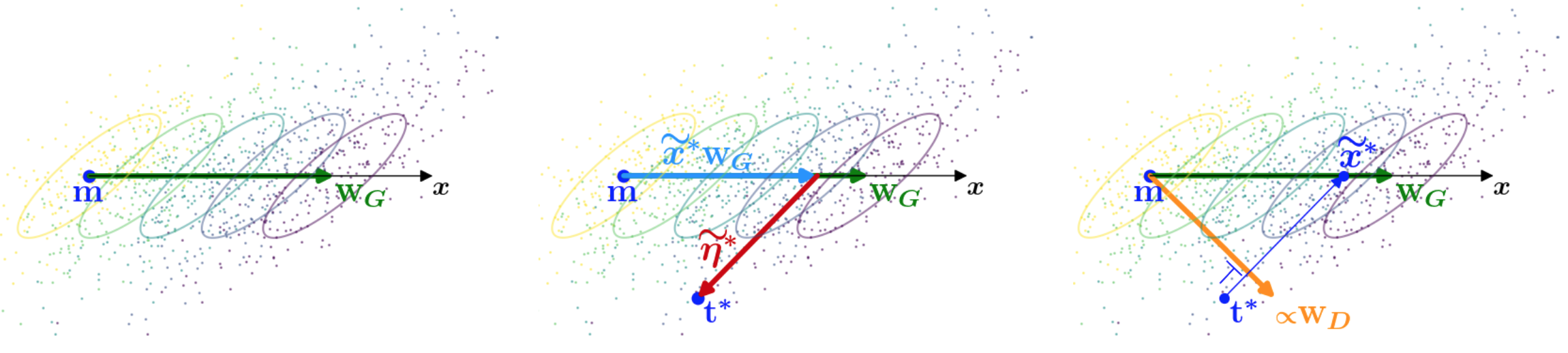}
  \caption{
 Left: 
 illustration of a the image generation process in a toy example.
 The figure shows the template $\fat{m}$, the generative map $\fat{w}_G$, and contour plots of the noise covariance matrix $\fat{C}$.
 Also shown are randomly generated 2D data points (dots in the figure),
 corresponding to 5 discrete values of a continuous variable of interest $x$ (marked with different colors). The discretization was performed for visualization purposes.
  %
  Middle: 
  when the model is inverted to compute a predicted value $\tilde{x}^*$ for a given data point $\fat{t}^*$ , the signal is effectively decomposed into $\fat{t}^*=\fat{m}+\tilde{x}^*\fat{w}_G+\tilde{\bldgr{\eta}}^*$.  
  Right: 
  the same result $\tilde{x}^*$ can also be obtained by 
  projecting the data point $\fat{t}^*$ orthogonally onto the direction $\fat{w}_D = \fat{C}^{-1} \fat{w}_G$. 
  This 
  operation is \textit{mathematically} equivalent to decomposing the signal into its components as in the middle figure, but not \textit{in terms of interpretability}, since causal effect and noise pattern are intermingled in $\fat{w}_D$.
  %
%
%
  }
  \label{fig:modelInversion}
\end{figure*}

The reason the noise covariance $\fat{C}$ is taken into account 
in $\fat{w}_D$
--
which in turn 
makes 
$\fat{w}_D$
hard to interpret
--
can be illustrated with a simple toy example involving only two ``voxels'', shown in Fig.~\ref{fig:modelInversion}. 
%
When 
the method is tasked with 
computing the prediction $\tilde{x}^*$ from an image $\fat{t}^*$, it effectively decomposes 
$\fat{t}^*$
into its most likely constituent components:
$\fat{m}$ (a population template),
 $\tilde{x}^*\fat{w}_G$ (the estimated causal effect),
and $\tilde{\bldgr{\eta}}^*$ (the most likely noise vector)
-- see \eqref{eq:decomposition} and Fig.~\ref{fig:generativeModelWithRealImages}.
%
In this process,
parts of the signal that are well-explained by the noise model $\fat{C}$ can be attributed to the noise component $\tilde{\bldgr{\eta}}^*$, and therefore effectively 
discarded
when estimating
$\tilde{x}^*$.
%
This is illustrated in Fig.~\ref{fig:modelInversion} (middle).
%
Mathematically, 
the same value $\tilde{x}^*$
can also 
be obtained
by simply computing the
inner product $\fat{w}_D^T\fat{t}^*$
(see \eqref{eq:mean} 
and Fig.~\ref{fig:modelInversionWithImages}),
which folds the process of 
separating the effect of interest from 
noise patterns
into a single operation.
This amounts to projecting the given data point $\fat{t}^*$ orthogonally onto the direction $\fat{w}_D$, as illustrated in Fig.~\ref{fig:modelInversion} (right).
However, 
this projection operation is hard to interpret
on its own:
$\fat{w}_D$ only reveals 
the final recipe of
how predictions are computed, but no longer the underlying logic.

\begin{figure*}[p]
  \setlength{\myWidth}{0.21\linewidth}
  \centering
  \begin{tabular}{cccc}
 \vspace{2mm}
  
   \textbf{47 years} &
   \textbf{64 years} &
   \textbf{80 year}s & 
   \\
    \includegraphics[align=c,
    clip,width=\myWidth]
    {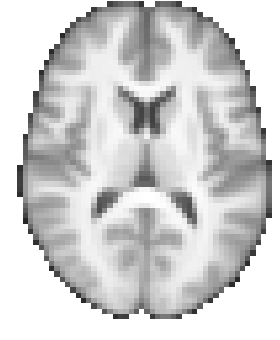}
    &
    \includegraphics[align=c,trim={0cm 0cm 0cm 0cm},clip,width=\myWidth]{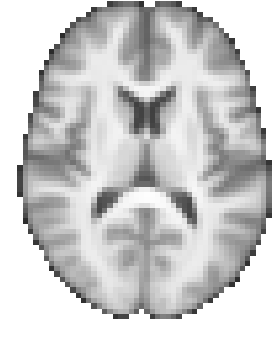}
    &
    \includegraphics[align=c,trim={0cm 0cm 0cm 0cm},clip,width=\myWidth]{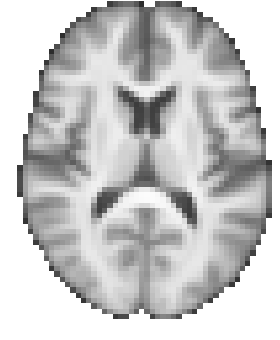}
    & 
      \includegraphics[align=c,trim={0cm 0cm 0cm 0cm},clip,width=0.065\linewidth]
    {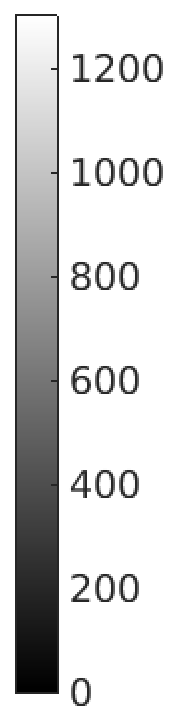}
    \\
  \end{tabular}  
  \caption{Age-specific templates 
  \revision{synthesized}
  by the proposed method trained on $N=2,600$ subjects, representing the expected ``average'' image for a specific age. Known age-related  effects, such as wider sulci and bigger ventricles~\citep{Fjell2009Cortex,fjell2010structural}, are clearly displayed.
  }
  \label{fig:ageSpecificTemplates}
\end{figure*}

\begin{figure*}[p]
   \setlength{\myWidth}{0.22\linewidth}
  \centering
  
  \begin{tabular}{cccc}
 \vspace{2mm}
  

    $\textbf{47}$ \textbf{years (real)} &
    \hspace{-9.00mm}
  $\textbf{80}$ \textbf{years (counterfactual)} &
   \hspace{3.00mm}
 $\textbf{80}$ \textbf{years (real)} &
    \hspace{-9.00mm}
  $\textbf{47}$ \textbf{years (counterfactual)}

   \\
     \includegraphics[trim={0cm 0cm 0cm 0cm},clip,width=\myWidth]
{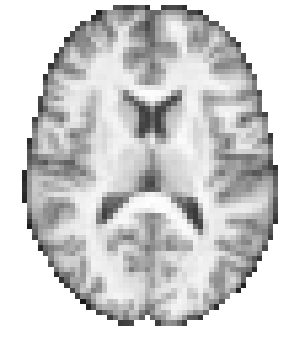}
    &
    \hspace{-8.00mm}
     \includegraphics[trim={0cm 0cm 0cm 0cm},clip,width=\myWidth]
{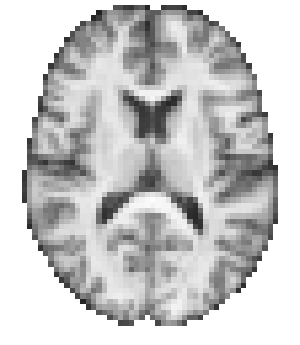}
    &
     \hspace{3.00mm}
     \includegraphics[trim={0cm 0cm 0cm 0cm},clip,width=\myWidth]
{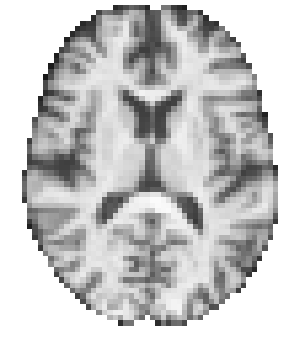}
    &
     \hspace{-8.00mm}
    \includegraphics[trim={0cm 0cm 0cm 0cm},clip,width=\myWidth]
{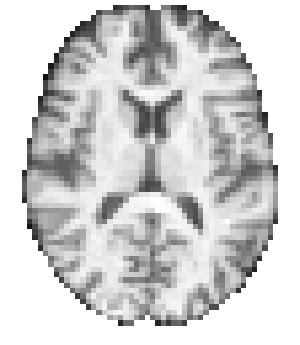}
  \end{tabular}  
  \caption{
  Two examples of counterfactuals 
  \revision{synthesized}
  by the model for age prediction, trained on $N=2,600$ subjects.  
  Left: real and counterfactual image of a 47-year old subject who is artificially aged to 80 years. Right: the same for a 80-year old subject who is rejuvenated to age 47.
  %
  }
  \label{fig:counterfactuals}
\end{figure*}

In addition to visualizing its generative maps directly for human interpretation, the causal interpretation of our model can also be used to simulate the effect of aging on actual images.
On the population level, this can be achieved by 
computing 
age-specific templates
$\fat{m} + x \fat{w}_G$ for different values of $x$,
as illustrated in Fig.~\ref{fig:ageSpecificTemplates}.
More detailed explanations can also be provided at the level of the individual subject,
using \emph{counterfactuals}~\citep{pearl2018book} --  
imaginary images of specific individuals if they had been younger or older.
Given an image $\fat{t}$ and the real age $x$, \eqref{eq:noiseVectorEstimation} can be used to compute 
the 
subject-specific
noise vector $\bldgr{\eta}$, which captures the subject's individual idiosyncrasies that are not explained by the population-level causal model.
Counterfactual images can then be obtained by 
re-assembling the forward model 
from its constituent components,
using a different, imaginary age $x$
in 
\eqref{eq:decomposition}.
Examples of this process are shown in Fig.~\ref{fig:counterfactuals}.

\revision{

The ability to generate synthetic images that are conditioned on the target variable $x$ 
provides 
alternative
visual explanations of the method's prediction process 
beside simply showing the 
generative map $\fat{w}_G$.
These 
will be
particularly
useful when the model is extended to include nonlinear effects
(so that the cause-effect relationship can no longer be described using a single 
linear
map, 
see Sec.~\ref{sec:nonlinearities}),
or when 
features are used that are more difficult to interpret than 
the voxel-level intensities used here
(e.g., parameters of a deformation field, as discussed in Sec.~\ref{sec:discussion}).
Counterfactual images in particular are thought to be closely aligned with 
human intuition%
~\citep{pearl2018book}: 
%
%
In a disease classification task
where a subject is predicted to suffer from a particular disease, 
for instance, 
a counterfactual would show 
their 
presumed brain shape 
in health 
for comparison,
mimicking how a human expert would explain 
their prediction to
others.

}

\myclearpage
\revision{%
\section{More Complex Variants}
\label{sec:extensions}
}

\noindent
Here we show how the 
\revision{basic form of the model described in Sec.~\ref{sec:method}}
can 
\revision{easily be}
extended to take into account additional subject-specific covariates \revision{and confounders}, as well as nonlinear dependencies on the variable of interest.

%
%
%

\bigskip
\revision{%
\subsection{Covariates and confounders}
\label{sec:covariates}
}

\noindent
When additional demographic or clinical 
information is available about the subjects,
it is straightforward to include this in the model
\revision{%
--
either as additional covariates (Fig.~\ref{fig:graphs}(b))
or as confounders (Fig.~\ref{fig:graphs}(c)).
}
This can be useful for further improving prediction accuracy, 
or for removing the effect of confounders that would otherwise invalidate the causal interpretation of the generative maps
\revision{as demonstrated below.}

Assuming each subject has $L$ extra variables $y^1, \ldots, y^L$, the model \eqref{eq:decomposition} can be extended to
\begin{equation*}
  \fat{t}=\fat{m} + x \fat{w}_G + \sum_{l=1}^L y^l \fat{w}_y^l + \bldgr{\eta},
\end{equation*}
where 
$\fat{w}_y^1, \ldots, \fat{w}_y^L$
are now 
extra spatial weight maps that also need to be estimated.
%
During training, the corresponding 
parameters
$\fat{W} = ( \fat{m}, \fat{w}_G, \fat{w}_y^1, \ldots, \fat{w}_y^L )$, $\fat{V}$ and $\bldgr{\Delta}$ can still be estimated using
\eqref{eq:W}, \eqref{eq:updateV} and \eqref{eq:updateDelta},
provided that 
$\bldgr{\phi}_n = (1, x_n, y_n^1, \ldots, y_n^L )^T$ is used instead of $\bldgr{\phi}_n = (1, x_n )^T$.
To predict an unknown variable of interest $x^*$ from a subject with image $\fat{t}^*$
and covariates $y^{*1}, \ldots, y^{*L}$,
\eqref{eq:prediction_classification} and \eqref{eq:mean} remain valid when
$\fat{t}^*$ is replaced by 
$\left( \fat{t}^* - \sum_l y^{*l}\fat{w}_y^{l}  \right)$.

To 
\revision{demonstrate the resulting model,}
we considered a classification experiment of 
%
%
%
\revision{%
Alzheimer's disease (AD) patients vs.~healthy controls. 
For this experiment, we used the OASIS-1 dataset\footnote{\url{https://www.oasis-brains.org}} (age range 18-96 years)
consisting of T1-weighted scans of 
100 AD subjects (average age $77 \pm 7.12$ years) and 336 controls (average age $44 \pm 24$ years).
}%
\revision{%
Importantly, 
we included age as a known 
confounder%
,
since
it both affects 
the imaging data
$\fat{t}$
directly and is a partial cause of the disease status $x$
(because AD occurs more frequently in older subjects, and the two disease groups are not age-matched in this dataset).
Because there are also more controls than AD subjects,
we used unequal priors of
0.7}\revisiontwo{7}\revision{ vs.~0.2}\revisiontwo{3}\revision{, respectively.
%
}

As input features we used 
probabilistic gray matter segmentations computed with SPM12, 
after warping them to 
standard space
and modulating them~\citep{ashburner2014spm12} to 
preserve 
signal
that would otherwise be removed by the spatial normalization.
%
Given the limited number of subjects in the dataset,
\revision{we determined the optimal value of the method's hyperparameter $K$ in}
a nested 5-fold cross-validation (CV) setting:
For each 
data 
split 
of the outer CV loop
into a training and a test set,
\revision{an internal 5-fold CV was performed within the training set,}
evaluating different values of $K$.
The 
\revision{value with the highest classification accuracy}
\revision{was}
then 
\revision{retained}
and 
used to re-train the model on the 
entire
training set,
which was then evaluated 
on the test set.
This procedure was repeated for each 
data split in the outer loop, to yield 
\revision{an overall prediction score}.

\begin{figure*}[t!]
  \centering
  \setlength{\myWidth}{0.30\linewidth}
  \setlength{\fboxrule}{2pt}
  \fcolorbox{white}{white}{
  \begin{tabular}{cccc}
    \centering
    \includegraphics[trim={0cm 0cm 0cm 0cm},clip,width=0.20\linewidth]{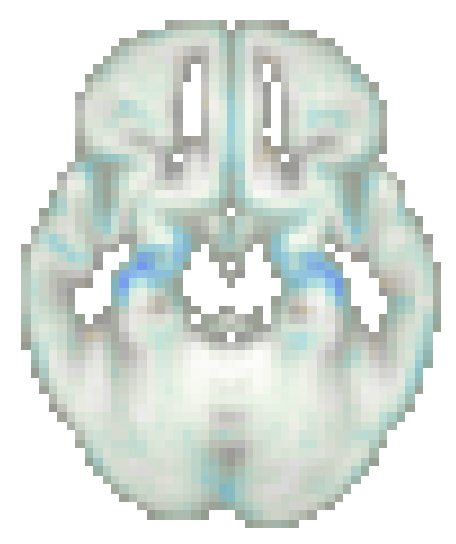} &
    \includegraphics[trim={0cm 0cm 0cm 0cm},clip,width=0.21\linewidth]{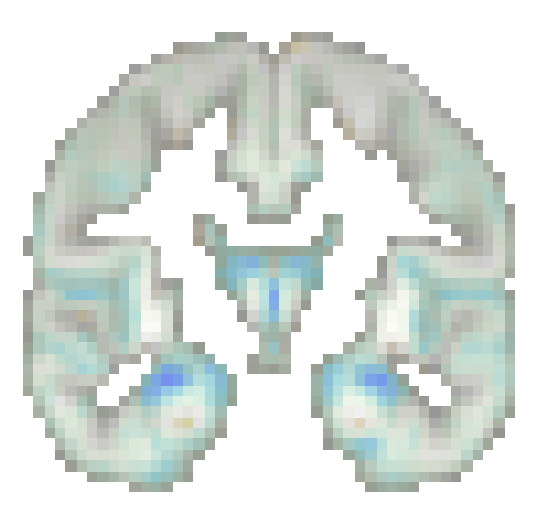} &
    \includegraphics[trim={0cm 0cm 0cm 0cm},clip,width=0.24\linewidth] {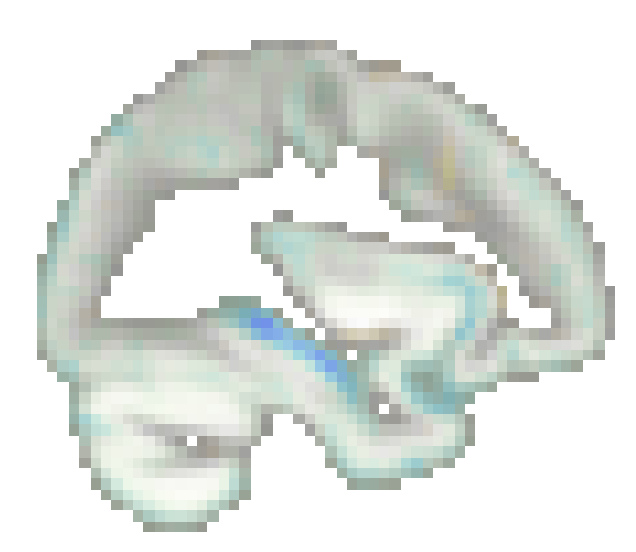} &
    \hspace{0.5mm}
        \includegraphics[trim={0cm 0cm 0cm 0cm},clip,width=0.055\linewidth]{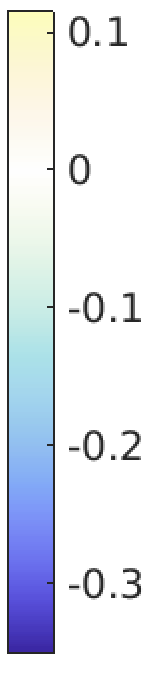} 
          \\
  \includegraphics[trim={0cm 0cm 0cm 0cm},clip,width=0.20\linewidth]{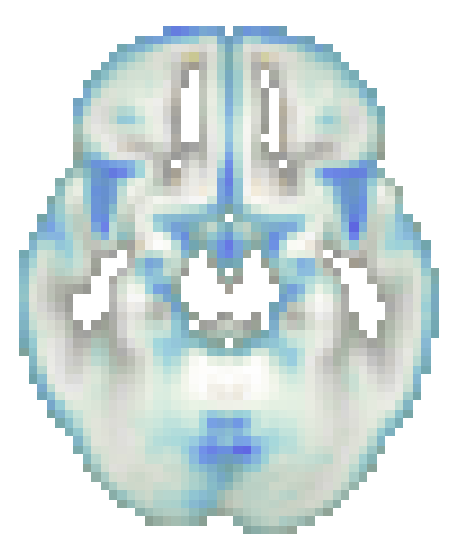} &
    \includegraphics[trim={0cm 0cm 0cm 0cm},clip,width=0.21\linewidth]{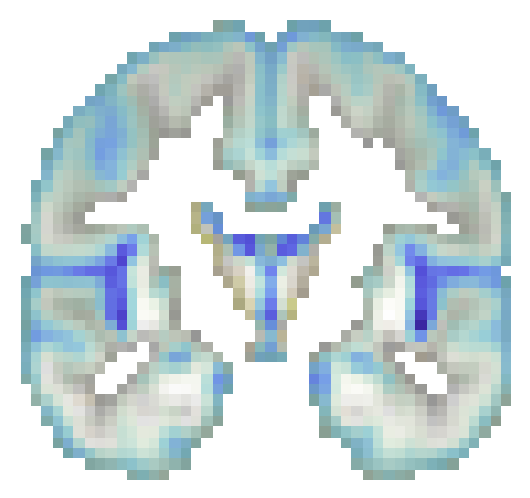} &
    \includegraphics[trim={0cm 0cm 0cm 0cm},clip,width=0.24\linewidth] {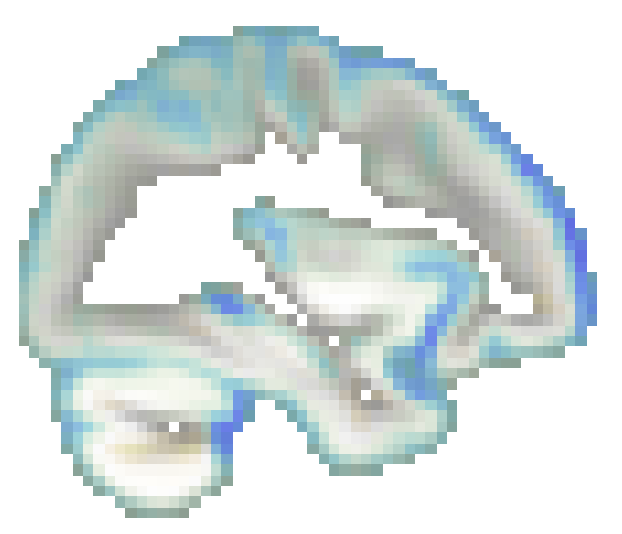} &
        \includegraphics[trim={0cm 0cm 0.3cm 0cm},clip,width=0.055\linewidth]{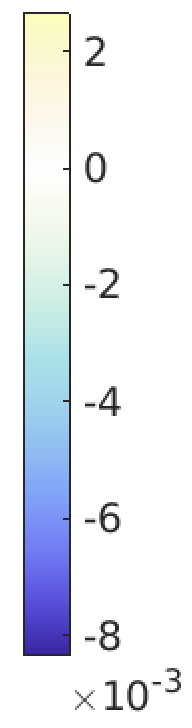}
      \end{tabular}  
      }
 \caption{
 \revision{%
  Generative maps obtained in a disease classification experiment based on gray matter segmentations, when age is included as a confounder:
  $\fat{w}_G$ (top) and $\fat{w}_y$ (bottom) express the effect of AD and age, respectively. The maps are overlaid on the 
  population 
  template.
  Voxels with zero weight are transparent.
  }
  }
  \label{fig:ad_and_age_maps_oasis}
\end{figure*}

\begin{figure*}[t!]
  \centering
  \setlength{\myWidth}{0.30\linewidth}
  \setlength{\fboxrule}{2pt}
  \fcolorbox{white}{white}{
  \begin{tabular}{cccc}
    \centering
    \includegraphics[trim={0cm 0cm 0cm 0cm},clip,width=0.20\linewidth]{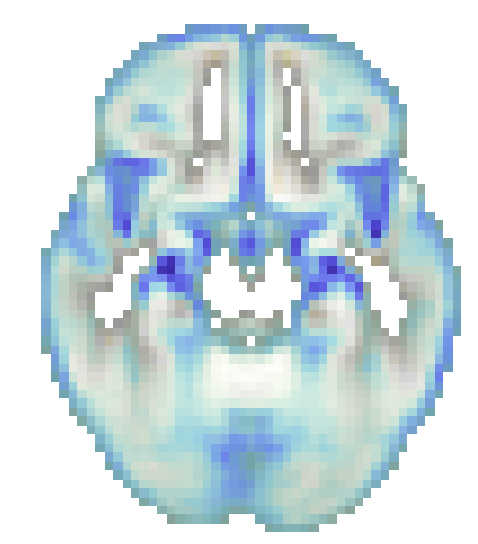} &
    \includegraphics[trim={0cm 0cm 0cm 0cm},clip,width=0.21\linewidth]{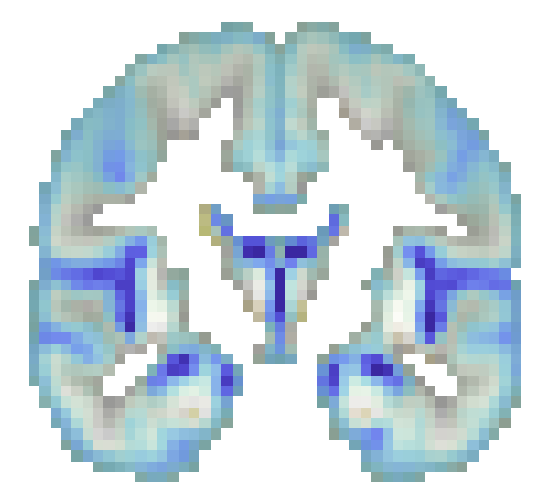} &
    \includegraphics[trim={0cm 0cm 0cm 0cm},clip,width=0.24\linewidth] {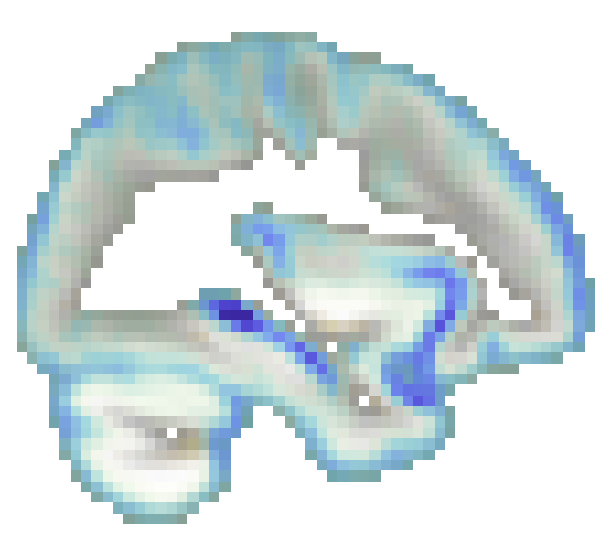} &
    \hspace{0.5mm}
        \includegraphics[trim={0cm 0cm 0cm 0cm},clip,width=0.055\linewidth]{colorbar_ad_maps_oasis.png} 
      \end{tabular}  
      }
 \caption{
 \revision{%
 Generative map $\fat{w}_G$ obtained in the same disease classification experiment shown of Fig.~\ref{fig:ad_and_age_maps_oasis}, but without controlling for age in the model. The map is now biased to also reflect the effect of aging instead of just the AD effect. 
  }
  }
  \label{fig:ad_without_age_maps_oasis}
\end{figure*}

%
\revision{%
We obtained an overall classification accuracy of 0.881.
Fig.~\ref{fig:ad_and_age_maps_oasis} shows the obtained spatial map $\fat{w}_G$ of the disease effect in one of the folds, 
together with the estimated age effect $\fat{w}_y$.
Since the confounding effect of age is automatically controlled for in the model, 
$\fat{w}_G$ 
reflects the average brain changes that occur in direct response to AD disease
(highlighting hippocampal atrophy in particular).
For comparison, we repeated the same experiment but this time \emph{without} including age as a confounder, simulating the setting of 
uncontrolled confounding shown in 
Fig.~\ref{fig:graphs}(d).
Although the model still predicts well in this setting (classification accuracy 0.874), the estimated ``disease effect'' $\fat{w}_G$ is now misleading as it 
it is biased to 
strongly
reflect the effect of aging (see Fig.~\ref{fig:ad_without_age_maps_oasis}).
The spatial map still indicates which image areas the method is paying attention to when making predictions, but now merely 
shows a statistical association rather than a direct causal relationship.


}

%
%

\bigskip
\subsection{Nonlinearities in the causal model}
\label{sec:nonlinearities}

\noindent
Another extension of~\eqref{eq:decomposition} is to consider nonlinear cause-effect relations in the model:
\begin{equation*}
  \fat{t}=\fat{m} + x \fat{w}_G + \sum_{l=1}^L f_l( x ) \fat{w}_y^l + \bldgr{\eta},
\end{equation*}
where each $f_l( x )$ is some nonlinear function of the (continuous) variable of interest $x$.
%
\revision{%
Although 
this corresponds to the simple causal diagram shown in Fig.~\ref{fig:graphs}(a),
during training it
}%
can be viewed as 
a special case of 
the model extension of Sec.~\ref{sec:covariates}: 
the 
mappings 
$f_l( \cdot ), \,\, l=1,\ldots,L$ 
can be evaluated for each training subject,
and treated as known additional covariates in the model.
Predicting with a trained model is no longer governed by the linear equation \eqref{eq:mean}, though.
We therefore invert the model
by finely discretizing $x^*$ into $P$ distinct values $x_{p}, p=1,\ldots,P$, and evaluating the posterior probability of each.
Assuming a flat prior, this yields
\begin{equation}
  p( x^* = x_p | \fat{t}^*, \fat{W}, \fat{C} )
  = 
  \frac{ 
    \gamma_p
    }
    {
    \sum_{p'=1}^P
   \gamma_{p'}
    }
    ,
  \nonumber
\end{equation}
where 
$\gamma_p = \mathcal{N}( \fat{t} | \fat{m} 
    + x_p \fat{w}_G 
    + \sum_{l=1}^L f_l(x_p) \fat{w}_y^l, 
    \fat{C} ) $
can be evaluated efficiently using~\eqref{eq:logML_singlesubject}, \eqref{eq:invC_trick} and \eqref{eq:detC_trick}.
Predictions are then obtained as the expected value of this posterior distribution:
$$
  \tilde{x}^*
  =
  \sum_{p=1}^P x_p p( x^* = x_p | \fat{t}^*, \fat{W}, \fat{C} )
  .
$$

In order to demonstrate this 
\revision{variant},
we tested whether the inclusion of an extra quadratic term in the forward model (i.e., $f_1(x) = x^2$) can improve age estimation results compared to the basic
linear model. 
For this purpose,
we used modulated gray matter segmentations (computed in the same way as described in~Sec.~\ref{sec:covariates}) of T1-weighted scans of 562 healthy subjects from the IXI dataset\footnote{\url{https://brain-development.org/ixi-dataset/}}.
%
This dataset was selected because aging has been shown to have an 
approximately quadratic 
effect across adulthood on some brain structures  
\citep{walhovd2005effects}, 
and the IXI dataset 
covers an age span 
that is large enough 
(20-86 years) 
to possibly 
exploit
this behavior
for prediction purposes.
%
%
\revision{%
In our 
experiment,
we treated the possible inclusion of the quadratic term 
as an extra binary hyperparameter, 
determined in the same way as the other hyperparameter $K$ of the method 
within a nested 5-fold CV (as in Sec.~\ref{sec:covariates}).
}%
%
%
%
For the discretization, we used $P=20$ intervals covering the entire age range.

\begin{figure*}[t!]
  \centering
  \setlength{\myWidth}{0.30\linewidth}
  \setlength{\fboxrule}{2pt}
  \fcolorbox{white}{white}{%
  \begin{subfigure}[b]{\linewidth}
  \begin{tabular}{ccccc}
    \centering
   
   $\textbf{20}$ \textbf{years} &
        $\textbf{40}$ \textbf{years} &
        $\textbf{60}$ \textbf{years} &
        $\textbf{80}$ \textbf{years}\\
  \includegraphics[trim={0cm 0cm 0cm 0cm},clip,width=0.20\linewidth]{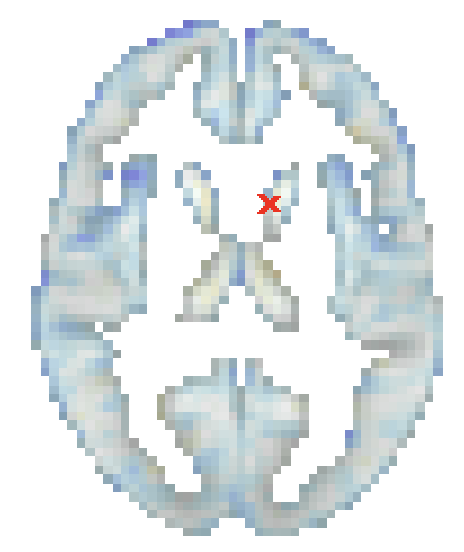} &
    \includegraphics[trim={0cm 0cm 0cm 0cm},clip,width=0.20\linewidth]{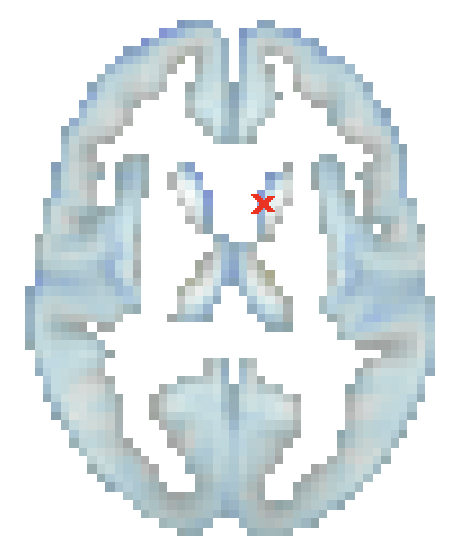} &
    \includegraphics[trim={0cm 0cm 0cm 0cm},clip,width=0.20\linewidth] {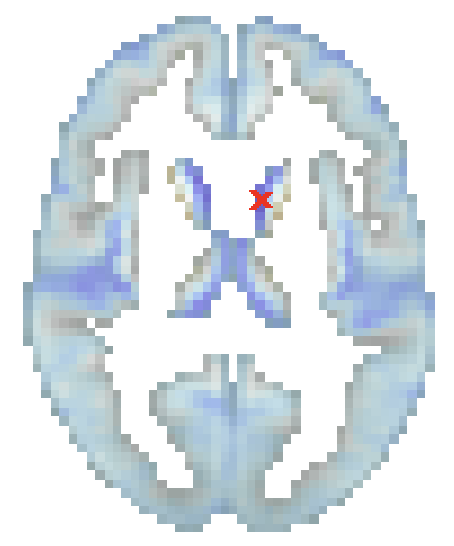} &
      \includegraphics[trim={0cm 0cm 0cm 0cm},clip,width=0.20\linewidth] {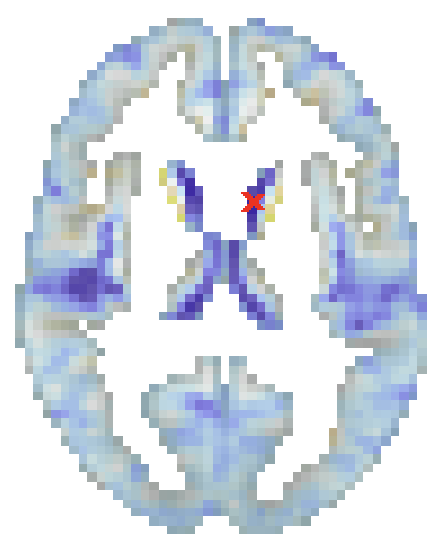} &
        \includegraphics[trim={0cm 0cm 0.3cm 0cm},clip,width=0.055\linewidth]{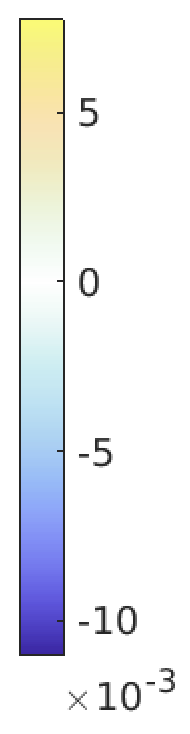}
        \end{tabular}  
 \caption{Quadratic model
  }
  \end{subfigure}
  }
  \setlength{\fboxrule}{2pt}
  \fcolorbox{white}{white}{%
  \begin{subfigure}[c]{0.45\textwidth}
   \begin{tabular}{cc}
    \centering
     \hspace{10.00mm}
    \includegraphics[trim={0cm 0cm 0cm 0cm},clip, width=0.45\linewidth]{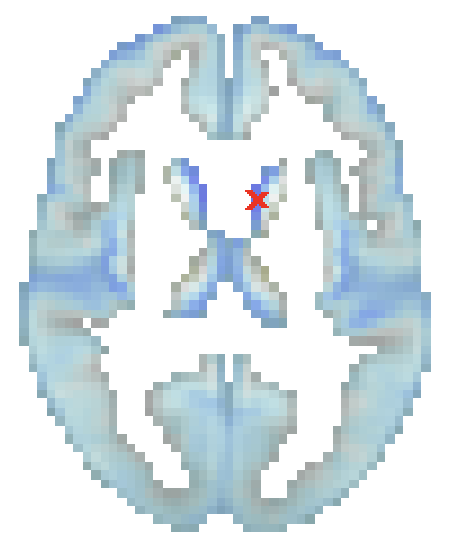} &
    \hspace{0.5mm}
        \includegraphics[trim={0cm 0.3cm 0cm 0cm},clip,width=0.14\linewidth]{colorbar_age_maps_ixi.png} 
          \\
      \end{tabular}  
 \caption{Linear model
  }
   \end{subfigure}
  }  
  \hfill
  ~
  \setlength{\fboxrule}{2pt}
  \fcolorbox{white}{white}{%
   \begin{subfigure}[c]{0.45\textwidth}
    \hspace{-2.00mm}
    \includegraphics[trim={0cm 0cm 0cm 0cm},clip,width=1.1\linewidth]
    {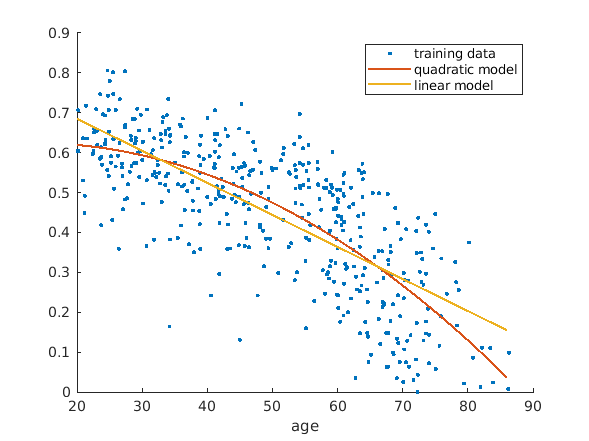} 
 \caption{Quadratic and linear fit in a selected voxel
  }
   \end{subfigure}
   }
   \caption{
   \revision{
   (a) Age-specific maps expressing the effect of aging, 
   overlaid on the population template,
   when a quadratic aging model is used.
   These maps were obtained by computing for each voxel the tangent to the quadratic model at the specified age.
  Voxels with zero weight are transparent.
  (b) Corresponding generative map $\fat{w}_G$ expressing the age effect obtained when a basic linear model is used instead.
  (c) Training data in a selected voxel (marked in red in figures (a) and (b)), together with the fitted quadratic and linear models.
  }
  }
  \label{fig:age_maps_ixi}
\end{figure*}

Using this set-up,
we found that the quadratic 
model was selected in all the CV folds,
and yielded 
\revision{significantly}
smaller age prediction test errors 
than the 
\revision{linear}
method
(MAE of 
4.36
vs.~%
4.73
years, \revision{p-value 
$<0.001$ with a one-sided paired t-test}).
%
\revision{Fig.~\ref{fig:age_maps_ixi} shows 
linearized maps demonstrating the effect of aging, evaluated as several 
age points,
for the quadratic model estimated in one of the folds (a),
together with the corresponding map $\fat{w}_{G}$ 
obtained with the linear model (b). 
We observe that the aging effect is more pronounced in elderly subjects, 
as is also clearly displayed by the curve fitted in a single selected voxel shown in Fig.~\ref{fig:age_maps_ixi}(c).
%
%
In nonlinear cases such as this, subject-specific counterfactuals -- such as the ones shown in Fig.~\ref{fig:counterfactuals} -- remain straightforward to compute with the model, and may provide more useful 
individualized
explanations than the group-level linearizations of Fig.~\ref{fig:age_maps_ixi}(a). 


}

\myclearpage
\revision{
\section{Discussion}
\label{sec:discussion}
}

\noteToSelf{
Summary of our method, stressing its advantages (incl. explainabiity, simplicity, speed (minutes for typical N sizes), accuracy, easy of incorporating covariates, causual stuff such as group-wise trends and subject-wise counterfactuals, \ldots)
}

\noindent
In this paper, we have proposed a lightweight method for image-based prediction that is inherently interpretable. It is based on classical human brain mapping techniques, but includes a multivariate noise model that yields accurate subject-level predictions when inverted. Despite its simplicity, the method predicts well in comparison with state-of-the art benchmarks, especially in typical training scenarios (those with no more than a few thousand training subjects) where more flexible techniques become prone to overfitting.

\noteToSelf{
In this paper we gave a causal interpretation to our generative model [ refer to group-level intervention (level 2 in Pearl's ladder) and subject-level counterfactuals (level 3) ], which is justified in our age, gender and disease status prediction experiments: changing age/gender/disease will change brain morphometry. But this interpretation requires that causual diagram [ if we have it in the paper ] is correct: (1) direction of causal effect is correct [ for example test score isn't causative, citep Wachinger paper ] and (2) no unobserved confounders [ explain what those are, maybe give example of MS where gender is controlled for even in the simple model by fiddling with the sample ]. In other scenarios method can still be used but generative maps only shows statistical associations, rahter than causations. [ Probably too long: Other alternative is to use other latent-variable models, closely related to ours, for example Wachinger and Asburner joint FA ]
}

\noteToSelf{
Method demonstrated/proposed here is basic and can be extended. Input data can be WM/GM/CSF, one hyperparam can be automatically inferred. Assumption of shared/identical covariance can be relaxed: for instance in classification can be independent [ resulting in QDA instead of LDA ] or scaled version [ if Friston has good paper we can cite ] allowing diseased subjects to have more variation [ or more general allow noise model to depend on target variable ]. Missing or irregular data: example of former is where training labels are not known with certainty [ e.g., AD diagnosis ], can use mixture of FAs using joint EM to simultaneously infer labels and estimate model parameters [ careful: can we? Don't conflate causal part with noise part! ]; example of latter is longitudinal data with missing/varying timings and number of visits.
}

The method 
we described here
represents a basic 
algorithm 
that 
can be further 
developed
with
more advanced
techniques.
%
For example,
while we used external cross-validation to determine a suitable number of latent variables 
in the noise model, 
methods
exist to infer this hyperparameter automatically from the training data itself~
\citep{bishop1998bayesian}.
%
It should also be 
possible
to address 
problems where the 
prediction
targets 
$x$
are not fully known
in the training data,
for instance 
%
by having the EM algorithm 
that 
currently
estimates
the 
noise covariance
also infer missing mixture class memberships~\citep{ghahramani1996algorithm}.
%
Examples of such scenarios include 
disease classification tasks in which the ``ground truth'' diagnosis is noisy, 
or 
semi-supervised learning tasks
in which 
a small 
training dataset 
is augmented with 
a large 
unlabeled one
to improve prediction performance~\citep{kingma2014semi}.
%
%
%
Finally,
the generative model can 
be 
further
extended to allow for more complex imaging data,
such as a combination of 
multimodal images~
\revision{\citep{liem2017predicting,engemann2020combining,cole2020multimodality},}
or longitudinal data where temporal correlations need to be explicitly 
taken into account~\citep{bernal2013statistical,bernal2013spatiotemporal}.

\revisiontwo{

This paper
only addressed \emph{encoding} scenarios
in which
an unknown condition $x$ is assumed to be the cause of changes observed in imaging data $\fat{t}$
(see Fig.~\ref{fig:graphs}(a)-(c)).
In this setting, 
it is well-known that 
only the forward model 
--
represented by the
generative maps $\fat{w}_G$ in the basic linear version 
of the proposed method
--
can be unambiguously interpreted,
whereas methods working in the inverse, anti-causal direction 
(to predict $x$ from $\fat{t}$)
cannot:
%
Their discriminative maps $\fat{w}_D$ 
may
include 
areas 
merely 
to 
remove 
variations in $\fat{t}$ that are not due to $x$,
while 
at the same time
missing other areas
affected by $x$ 
that
do not help in the prediction process
%
%
(interpretation rules S7 and S4 in~\citep{weichwald2015causal}, respectively).
The contribution of this paper is therefore to provide users with access to 
a
causal 
forward model, so that intuitive explanations of the resulting prediction procedure can be generated.
%
%
%
%
Unfortunately, 
this framework
is not applicable 
in \emph{decoding} scenarios 
where $\fat{t}$ causes $x$ rather than vice versa
(see Fig.~\ref{fig:graphs}(e)),
excluding
many 
important 
applications
aimed at understanding how brain-based abnormalities give rise to symptoms.
%
%
This is because the interpretation of such 
models works very differently:
for instance, 
while $\fat{w}_D$ remains difficult to interpret directly (rules R3 and R4 in~\citep{weichwald2015causal}), 
areas highlighted in $\fat{w}_G$ 
no longer
necessarily 
imply a causal relationship with $x$
(rule R1).

Although we have used causal language to 
motivate a generative modeling approach to subject-level prediction,
the 
generative maps depicted in this paper should not be over-interpreted as 
representing
unbiased estimates of
causality inferred 
from observational imaging data.
Due to various measurement and selection biases, each model will necessarily be 
specific 
only to the dataset it was derived from, 
without 
generalizing to other datasets the way a 
population-level unbiased model
would~\citep{wachinger2021detect}.
For instance, the UK Biobank cohort is known to not be representative of the general population with regard to a number of sociodemographic, physical, lifestyle, and health-related characteristics~\citep{fry2017comparison},
and this will be reflected in the generative maps that are produced.
Similarly,
there is a strong dependency of MRI-derived features on such factors as 
the scanners, the pulse sequences and the image processing pipelines that are used,
which will also find their way into the model estimates.
Finally, a key assumption 
to preserve interpretability 
in the proposed model 
is that all confounders are known,
so that the situation depicted in Fig.~\ref{fig:graphs}(d) can be excluded.
Although the requirement of affecting both 
a 
patient's condition
and 
their 
imaging data
independently dramatically reduces the number of potential confounders
(for instance, of the hundreds of potential confounds studied for UK Biobank brain imaging~\citep{alfaro2021confound}, very few beyond age, gender and perhaps imaging site would likely qualify),
we may not know all 
relevant factors nor have data on them in practice, resulting in confounding bias~\citep{wachinger2021detect}.

}

\revision{

Although this paper focused on interpretable 
\emph{models}
for prediction,
it is worth remembering that 
the interpretability of the resulting 
system
also 
depends 
on the 
type of 
\emph{features} that 
are 
used.
%
%
%
In the examples given throughout 
this
paper, the features 
consisted of 
voxel-level 
intensities
obtained after warping each subject into a common template space with nonlinear registration. 
While 
nonlinear 
registration
helps 
remove irrelevant inter-subject shape variations from the data,
it 
can also hide 
biological effects:
In the age-conditioned 
synthesized images 
of
Fig.~\ref{fig:ageSpecificTemplates} and~\ref{fig:counterfactuals}, for instance, a significant portion of 
the
atrophy patterns
caused by aging
(namely the part encoded in the nonlinear warps)
is missing.
%
%
%
This could potentially be avoided by 
using
the parameters governing 
the warps
-- such as the stationary velocity fields in diffeomorphic registration models~\citep{ashburner2007fast} -- 
as 
(additional)
features in the proposed model. 
%

%
A related issue is how much of a prediction method's robustness to inter-subject variability 
should be the responsibility of data preprocessing vs.~that of the prediction model itself.
%
%
It could be argued that, in our experiments, 
much of the ``heavy lifting'' was done by 
the
preprocessing
so that even simple linear-Gaussian models could work well.
%
This point, however, is nuanced by the fact that 
some form of
preprocessing is currently required for all subject-level prediction methods
in the literature.
In the SFCN paper, for instance, it is demonstrated that 
a neural network 
can
``model away'' nonlinear deformations that have not been removed from the input images;
however 
the images have 
still
been 
preprocessed with 
affine registration and
a 
full 
segmentation pipeline 
(paradoxally involving nonlinear registration)
for
skull stripping, bias field correction and intensity normalization.
%
%

%
The 
question of
data preprocessing
vs.~modeling
is further 
complicated
by the issue of scanner- and sequence-dependent MRI contrast in real-world 
applications: 
A 
subject-level prediction 
method that is trained on raw intensities will rarely be 
directly
applicable to data acquired elsewhere.
In principle, 
all these issues
can be addressed 
by integrating the type of 
forward models used in 
Bayesian 
segmentation%
~\citep{ashburner2005unified,puonti2016fast} 
-- 
which include 
subject-specific template deformation, bias field correction and 
contrast-adaptive intensity modeling
--
within the proposed generative approach itself.
%
Whether 
this 
will yield
a tangible benefit compared to 
preprocessing the data 
depends 
on the difficulty of training 
and inverting
such integrated models
in practice.
%
As demonstrated by the VAE example analyzed in this paper, if building higher-capacity models requires 
making approximations to keep computations feasible,
prediction performance may be hurt 
rather than helped.

}

\iftrue
%
%
In conclusion,
we have proposed a lightweight generative model that 
is inherently interpretable
and that can still make accurate predictions.
Since it can easily be extended, we hope it will form a useful basis for future developments in the field.
%
%
\fi

\bigskip

\section*{Acknowledgments}

\noindent
This research was conducted using the UK Biobank Resource under Application Number 65657, and it was made possible in part by the computational hardware generously provided by the Massachusetts Life Sciences Center (https://www.masslifesciences.com/).
This work was supported by the National Institute of Neurological Disorders and Stroke under grant number R01NS112161. OP is supported by the Lundbeck Foundation (grant R360-2021-395).


\newpage

 \appendix

\section{Derivations of model inversion}
\label{sec:appendix_predictions}

\noindent
Here we derive the expressions for making predictions about the variable of interest. %
%
For a binary target variable $x^*$, 
\begin{eqnarray*}
  & &
  \hspace{-8ex}
  p(x^*\!\!=\!\!1 | \fat{t}^*, \fat{W}, \fat{C} )  
  \\
  & = &
  \frac{p( \fat{t}^* | x^*\!\!=\!\!1, \fat{W}, \fat{C} ) p( x^*\!\!=\!\!1 ) }
       {p( \fat{t}^* | x^*\!\!=\!\!1, \fat{W}, \fat{C} ) p( x^*\!\!=\!\!1 ) + 
        p( \fat{t}^* | x^*\!\!=\!\!0, \fat{W}, \fat{C} ) p( x^*\!\!=\!\!0 ) }
  \\    
  & = &
  \frac{1}{1 + \frac{p( \bm{t}^* | x^*=0, \bm{W}, \bm{C} ) p( x^*=0 )}
                    {p( \bm{t}^* | x^*=1, \bm{W}, \bm{C} ) p( x^*=1 )}}
  \\
  & = &
  \sigma\Big[ \log p( \fat{t}^* | x^*\!\!=\!\!1, \fat{W}, \fat{C} ) - \log p( \fat{t}^* | x^*\!\!=\!\!0, \fat{W}, \fat{C} ) 
  \\ 
  & & 
  + \log p(x^*\!\!=\!\!1) - \log p (x^*\!\!=\!\!0) \Big],
\end{eqnarray*}
where
\begin{eqnarray*}
  &  &
  \hspace{-8ex}
  \log p( \fat{t}^* | x^*\!\!=\!\!1, \fat{W}, \fat{C} ) - \log p( \fat{t}^* | x^*\!\!=\!\!0, \fat{W}, \fat{C} )
  \\
  & = &
  -\frac{1}{2}( \fat{t}^* - \fat{m} - \fat{w}_G )^T \fat{C}^{-1} ( \fat{t}^* - \fat{m} - \fat{w}_G )
  \\
  & &
  + \frac{1}{2}( \fat{t}^* - \fat{m} )^T \fat{C}^{-1} ( \fat{t}^* - \fat{m} )
  \\
  & = &
  \fat{w}_G^T \fat{C}^{-1} ( \fat{t}^* - \fat{m} ) -\frac{1}{2}\fat{w}_G^T \fat{C}^{-1} \fat{w}_G
  .
\end{eqnarray*}
This explains \eqref{eq:posterior_classification}. 

For a continuous target variable with flat prior,
the log-posterior
\begin{eqnarray*}
  & &
  \hspace{-8ex}
  \log p(x^* | \fat{t}^*, \fat{W}, \fat{C} )  
  \\
  & = &
  -\frac{1}{2}( \fat{t}^* - \fat{m} - x^* \fat{w}_G )^T \fat{C}^{-1} ( \fat{t}^* - \fat{m} - x^* \fat{w}_G )
  + 
  \mathrm{const}
\end{eqnarray*}
is quadratic with derivative
\begin{equation}
  \frac{d \log p(x^* | \fat{t}^*, \fat{W}, \fat{C} ) }{d x^*}
  =
  \fat{w}_G^T \fat{C}^{-1} ( \fat{t}^* - \fat{m} - x^* \fat{w}_G ) 
  \label{eq:derivative}
\end{equation}
and curvature
$$
  \frac{d^2 \log p(x^* | \fat{t}^*, \fat{W}, \fat{C} ) }{d {x^*}^2}
  = 
  -\fat{w}_G^T \fat{C}^{-1} \fat{w}_G
  .
$$
Therefore, the posterior is Gaussian, with variance given by \eqref{eq:variance}.
The mean is obtained by setting \eqref{eq:derivative} to zero, which yields 
\eqref{eq:mean}.

\section{Expression for $\fat{W}$}
\label{sec:appendix_W}

\noindent
For training,
the log marginal likelihood is given by 
\begin{eqnarray*}
  & &
  \hspace{-8ex}
  \log p\left( \{ \fat{t}_n \}_{n=1}^N | \{ x_n \}_{n=1}^N, \fat{W}, \fat{C} \right) 
  \\
  & = &
  \sum_{n=1}^{N} 
  -\frac{1}{2}
  (\fat{t}_n - \fat{W} \bldgr{\phi}_n)^T
  \fat{C}^{-1}
  (\fat{t}_n - \fat{W} \bldgr{\phi}_n )
  +
  \mathrm{const}
  ,
\end{eqnarray*}
which has as gradient with respect to $\fat{W}$:
$$
  \sum_{n=1}^N
  \fat{C}^{-1} (\fat{t}_n - \fat{W} \bldgr{\phi}_n ) \bldgr{\phi}_n^T
  .
$$
Setting to zero and re-arranging yields \eqref{eq:W}.

\section{Efficient implementation}
\label{sec:appendix_implementation}

\noindent
Using Woodbury's identity, we obtain
\begin{eqnarray}
  \fat{C}^{-1} 
  & = & 
  \bldgr{\Delta}^{-1} - \bldgr{\Delta}^{-1} \fat{V} 
  \left(
  \mathbb{I}_K + \fat{V}^T \bldgr{\Delta}^{-1} \fat{V}
  \right)^{-1}
  \fat{V}^T \bldgr{\Delta}^{-1}
  \nonumber \\
  & = & 
  \bldgr{\Delta}^{-1} - \bldgr{\Delta}^{-1} \fat{V} 
  \bldgr{\Sigma}
  \fat{V}^T \bldgr{\Delta}^{-1}
  \label{eq:invC}
  ,
\end{eqnarray}
and therefore
\eqref{eq:discriminativeWeights}
can be computed as
$$
\fat{w}_D 
= 
\bldgr{\Delta}^{-1}
\fat{w}_G
- \bldgr{\Delta}^{-1} \fat{V} 
  \bldgr{\Sigma}
  \left(
  \fat{V}^T \bldgr{\Delta}^{-1} 
  \fat{w}_G
  \right)
  .
$$
Using this result, \eqref{eq:variance} is given by
$
  v
  = 1 / ( \fat{w}_D^T \fat{w}_G ) 
  .
$

Computing 
the marginal likelihood \eqref{eq:marginalLikelihood}
--
which is 
needed 
both
to monitor convergence of the EM algorithm during model training,
and to invert the model with nonlinearities (Sec.~\ref{sec:nonlinearities}) --
involves numerical evaluations of the form:
\begin{equation}\label{eq:logML_singlesubject}
\log
\mathcal{N}\left( \bldgr{\eta} |~ \fat{0},\fat{C}\right)
\,\,
\propto
\,\,
\bldgr{\eta}^T \fat{C}^{-1} \bldgr{\eta}
+
\log
| \fat{C} | 
+
\mathrm{const}
.
\end{equation}
Using \eqref{eq:invC}, the first term can be computed as
\begin{eqnarray}\label{eq:invC_trick}
\bldgr{\eta}^T \fat{C}^{-1} \bldgr{\eta}
& = & 
\bldgr{\eta}^T 
\bldgr{\Delta}^{-1} 
(
\bldgr{\eta}
- \fat{V} 
  \bldgr{\Sigma}
  \fat{V}^T \bldgr{\Delta}^{-1}
  \bldgr{\eta}
) 
\nonumber \\
& = &
\bldgr{\eta}^T 
\bldgr{\Delta}^{-1} 
(
\bldgr{\eta}
- \fat{V} 
  \bldgr{\mu}
),
\end{eqnarray}
with
$
\bldgr{\mu} =
\bldgr{\Sigma}
  \fat{V}^T \bldgr{\Delta}^{-1}
  \bldgr{\eta}
$ being an estimate of the latent variables.
The second term can be computed using Sylvester's determinant identity \citep{sylvester}:
\begin{eqnarray}
  |\fat{V} \fat{V}^T\bldgr{\Delta}^{-1}+\mathbb{I}_J| 
  & = &
  |\fat{V}^T\bldgr{\Delta}^{-1}\fat{V}+\mathbb{I}_K|
  \nonumber \\
  & = &
  |\bldgr{\Sigma}|^{-1}
  \nonumber
  ,
\end{eqnarray}
so that 
\begin{equation}\label{eq:detC_trick}
\log |\fat{C}| =
\log | \bldgr{\Delta} | 
-
\log | \bldgr{\Sigma} |
.
\end{equation}


Finally,
the EM update \eqref{eq:updateDelta} of the diagonal matrix $\bldgr{\Delta}$
can 
be computed one element at a time:
the diagonal element corresponding to the $j^{th}$ voxel is given by
\begin{eqnarray*}
\bldgr{\Delta}_{jj}
& = &
\frac{1}{N}
\fat{r}_j^T 
\left(
\fat{r}_j
- 
\fat{M} 
\fat{v}_j
\right)
\\
& = &
\frac{1}{N}
\| \fat{r}_j - \fat{M}\fat{v}_j \|^2
+ 
\frac{1}{N}
\fat{r}_j^T \fat{M} \fat{v}_j
-
\frac{1}{N}
\fat{v}_j^T \fat{M}^T \fat{M} \fat{v}_j
\\
& = &
\frac{1}{N}
\| \fat{r}_j - \fat{M}\fat{v}_j \|^2
+ 
\frac{1}{N}
\fat{v}_j^T
\left( \fat{M}^T \fat{M} + N \bldgr{\Sigma} \right)
\fat{v}_j
\\
& & \hspace{25ex}
-
\frac{1}{N}
\fat{v}_j^T \fat{M}^T \fat{M} \fat{v}_j
\\
& = &
\frac{1}{N}
\| \fat{r}_j - \fat{M}\fat{v}_j \|^2
+ 
\fat{v}_j^T \bldgr{\Sigma} \fat{v}_j
.
\end{eqnarray*}
Here 
$\fat{v}_j^T$ is the $j^{th}$ row of $\fat{V}$, 
$
\fat{M}
=
\left(
\bldgr{\mu}_1, \ldots, \bldgr{\mu}_N
\right)^T
$,
and
$
\fat{r}_j = 
( \eta_1^{~j}, \ldots, \eta_N^{~j} )^T
$
where
$\eta_n^{~j}$ denotes the $j^{th}$ element of 
$\bldgr{\eta}_n$.
The next to last step makes use of the fact that (see \eqref{eq:updateV})
$$
\fat{v}_j = \left( \fat{M}^T \fat{M} + N \bldgr{\Sigma} \right)^{-1}
\fat{M}^T \fat{r}_j
.
$$

\revision{%
\section{Analysis of the Haufe transformation}
\label{sec:appendix_haufe}
}

\noindent
When the target variable $x$ has zero mean in the training data, i.e., 
$(\sum_{n=1}^N x_n)/N = 0$, 
the solution of
~\eqref{eq:W} is given by
\begin{equation*}
\fat{m} = \frac{\sum_{n=1}^N \fat{t}_n}{N}
\end{equation*}
and 
\begin{equation}
\fat{w}_G
=  
\left(\sum_{n=1}^{N} \fat{t}_n x_n\right) \left( \sum_{n=1}^{N} x_n^2 \right)^{-1}
,
\label{eq:wG}
\end{equation}
because 
$$
\sum_{n=1}^{N} \bldgr{\phi}_n \bldgr{\phi}_n^T
=
\left(
\begin{array}{cc}
N & 0 \\
0 & \sum_{n=1}^{N} x_n^2
\end{array}
\right)
.
$$
Using the notation 
$
\fat{T}
=
\left(
\fat{t}_1, \ldots, \fat{t}_N
\right)
$
and
$
\fat{x} = (x_1, \ldots, x_n)^T
$,
~\eqref{eq:wG}
can also be written as 
\begin{equation}
\fat{w}_G = \fat{T} \fat{x} (\fat{x}^T \fat{x})^{-1}
.
\label{eq:wG_matrixForm}
\end{equation}
%

\begin{figure}[t!]
  \setlength{\myWidth}{0.4\linewidth}
  \centering
  \setlength{\fboxrule}{2pt}
  \fcolorbox{white}{white}{
  \begin{tabular}{@{}c@{}c@{}c@{}}
    \includegraphics[align=c,clip,width=\myWidth]{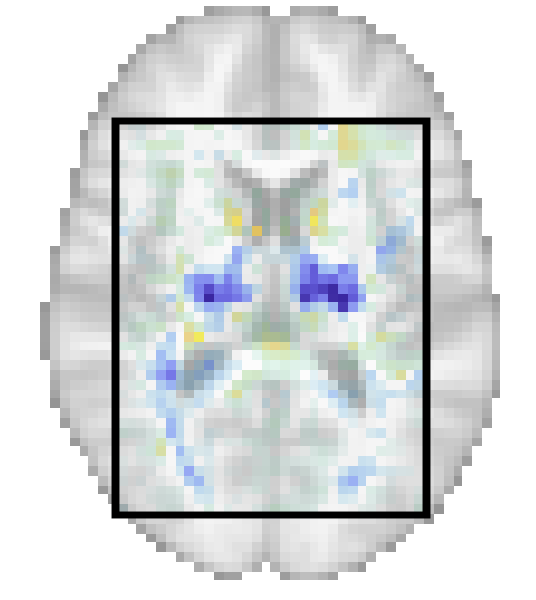} 
    &
    \includegraphics[align=c,clip,width=\myWidth]{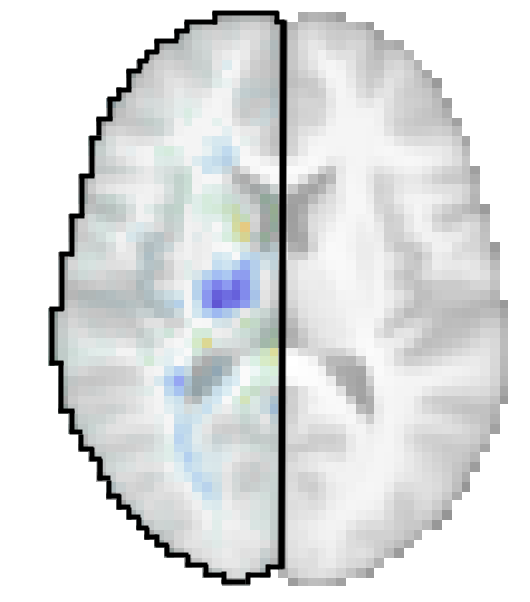} 
    & 
    \hspace{1.00mm}
    \includegraphics[align=c,clip,width=0.12\linewidth]{rvoxm_vertical_colorbar.png}
    \\
    \includegraphics[align=c,clip,width=\myWidth]{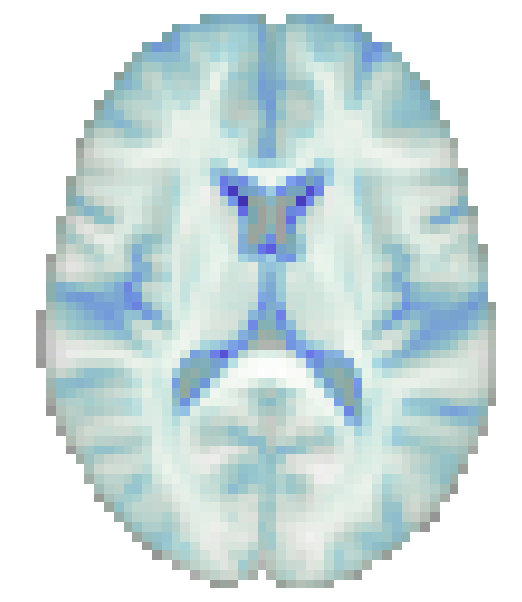} 
    &
    \includegraphics[align=c,clip,width=\myWidth]{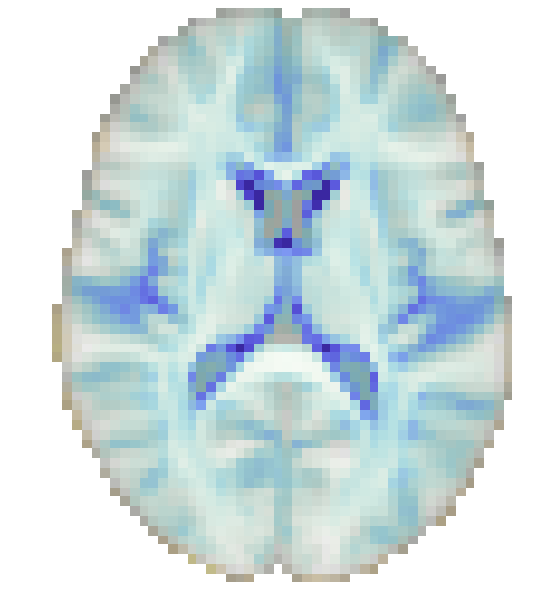} 
    & 
    \includegraphics[align=c,clip,width=0.1\linewidth]{wG_vertical_colorbar.png}
  \end{tabular}  
  }
  \caption{\revision{%
  Illustration of why the Haufe transformation 
  should not be used to explain the prediction process of a discriminative method.
  %
  %
  %
  Top: weights obtained when RVoxM is trained to predict age from $N=2,600$ 
  subjects when the weights are clamped to zero outside of two different ROIs indicated by the black outline. On the left the ROI encompasses only a cuboid area around the ventricles, whereas on the right only one hemisphere is included.
  Bottom: the corresponding maps $\fat{\tilde{w}}_G$ obtained with Haufe's 
  transformation
  hide the fact that voxels outside of the respective ROIs are never used in the predictions.
}
}
  \label{fig:haufe}
\end{figure}

\revision{
Given some linear discriminative method 
with
weight vector $\fat{w}$,
the method in~\citep{haufe2014interpretation} aims to recover a corresponding generative weight vector $\fat{\tilde{w}}_G$ 
as follows.
Rather than using the available \emph{real} training targets $\{x_n\}$
-- which were used to obtain $\fat{w}$ --
~\eqref{eq:wG_matrixForm} is instead applied to \emph{estimated} targets $\{\tilde{x}_n\}$ 
obtained with the discriminative method:
$$
\tilde{x}_n = \fat{w}^T \fat{t}_n, \quad \forall n,
$$
where it is assumed that the training images 
have been preprocessed to
have zero mean
($\fat{m} = \fat{0}$).
Writing $\fat{\tilde{x}} = \fat{T}^T \fat{w}$
and plugging this 
into
~\eqref{eq:wG_matrixForm}
then yields
$$
\fat{\tilde{w}}_G 
\, = \,
\fat{T} \fat{\tilde{x}} (\fat{\tilde{x}}^T \fat{\tilde{x}})^{-1} \\
\, = \,
\fat{T} ( \fat{T}^T \fat{w} ) (\fat{\tilde{x}}^T \fat{\tilde{x}})^{-1} \\
\, = \, 
\bldgr{\Sigma}_t
\fat{w}
\tilde{\sigma}_x^{-2}
$$
where
$
\bldgr{\Sigma}_t = N^{-1} \fat{T} \fat{T}^T
$
and 
$
\tilde{\sigma}_x^2 
= N^{-1} \fat{\tilde{x}}^T \fat{\tilde{x}} 
$.
This corresponds to
~(6) in~\citep{haufe2014interpretation}.
%

%
A fundamental
issue with 
the 
obtained
expression
$$
\fat{\tilde{w}}_G = \bldgr{\Sigma}_t \fat{w} \tilde{\sigma}_x^{-2}
$$
(the Haufe transformation)
is that it obscures the fact that the dependency of $\fat{\tilde{w}}_G$ on $\fat{w}$ is only through the predictions 
$\fat{\tilde{x}} = \fat{T}^T \fat{w}$. This is a problem, because good predictions can be obtained with very different $\fat{w}$'s -- including those ignoring entire image areas --
which prevents $\fat{\tilde{w}}_G$ from being a reliable indication of feature importance.
%
To illustrate this point, in Fig.~\ref{fig:haufe} we trained two different 
$\fat{w}$'s
to predict age in the UK Biobank dataset used in Sec.~\ref{sec:age_gender_prediciton} with the RVoxM method, while clamping the weights 
$w_j$ to zero in large (and very different) image areas that are therefore 
necessarily ignored when making predictions. 
Since both regimes still predict 
well (MAE within 6\% of what is obtained with RVoxM without clamping),
 the resulting $\fat{\tilde{w}}_G$'s 
look very similar to each other 
and to the corresponding map $\fat{w}_G$ shown in Fig.~\ref{fig:wGs}(middle), 
completely failing to reveal that certain image areas are never 
actually
looked at.
%


\bigskip
%

}

\bibliography{bibliography}






\if\includeOldVersions1
  \include{oldStuff}
\fi

\end{document}